%% file: main.tex
\documentclass[sigconf]{acmart}

\usepackage[utf8]{inputenc} 
\usepackage[T1]{fontenc}    
\usepackage{hyperref}       
\usepackage{url}            
\usepackage{booktabs}       
\usepackage{amsfonts}       
\usepackage{nicefrac}       
\usepackage{microtype}      
\usepackage{xcolor}         
\usepackage{multirow}       
\usepackage{graphicx}
\usepackage{cite}
\usepackage{amsmath}
\usepackage{amsthm}

\usepackage{appendix}
\usepackage{bbm}
\usepackage{bm}
\usepackage{mathrsfs}
\usepackage{enumitem}
\usepackage{mathtools} 
\usepackage{tikz} 
\usepackage{fancyhdr}
\usepackage[ruled,linesnumbered]{algorithm2e}
\usepackage{graphicx}

\AtBeginDocument{%
  \providecommand\BibTeX{{%
    \normalfont B\kern-0.5em{\scshape i\kern-0.25em b}\kern-0.8em\TeX}}}

\copyrightyear{2024}
\acmYear{2024}
\setcopyright{acmlicensed}\acmConference[KDD '24]{Proceedings of the 30th
ACM SIGKDD Conference on Knowledge Discovery and Data Mining}{August
25--29, 2024}{Barcelona, Spain}
\acmBooktitle{Proceedings of the 30th ACM SIGKDD Conference on Knowledge
Discovery and Data Mining (KDD '24), August 25--29, 2024, Barcelona, Spain}
\acmDOI{10.1145/3637528.3671516}
\acmISBN{979-8-4007-0490-1/24/08}

\acmConference[KDD'24]{ACM SIGKDD Conference on Knowledge Discovery and Data Mining}{August 25--29, 2024}{Barcelona, Spain}
\begin{document}

\title{Rankability-enhanced Revenue Uplift Modeling Framework for Online Marketing}


\author{Bowei He}
\authornote{Work done as an intern in FiT, Tencent}
\affiliation{%
  \institution{City University of Hong Kong}
  \country{Hong Kong SAR}
}
\email{boweihe2-c@my.cityu.edu.hk}

\author{Yunpeng Weng}
\affiliation{%
  \institution{FiT, Tencent}
  \city{Shenzhen}
  \country{China}
  }
\email{wengyp@mail3.sysu.edu.cn}

\author{Xing Tang}
\affiliation{%
  \institution{FiT, Tencent}
  \city{Shenzhen}
  \country{China}
  }
\email{xing.tang@hotmail.com}

\author{Ziqiang Cui}
\affiliation{%
  \institution{City University of Hong Kong}
  \country{Hong Kong SAR}
}
\email{ziqiang.cui@my.cityu.edu.hk}

\author{Zexu Sun}
\affiliation{%
  \institution{Renmin University of China}
  \city{Beijing}
  \country{China}
}
\email{sunzexu21@ruc.edu.cn}

\author{Liang Chen}
\affiliation{%
  \institution{FiT, Tencent}
  \city{Shenzhen}
  \country{China}
  }
\email{leocchen@tencent.com}

\author{Xiuqiang He}
\affiliation{%
  \institution{FiT, Tencent}
  \city{Shenzhen}
  \country{China}
  }
\email{xiuqianghe@tencent.com}

\author{Chen Ma}
\authornote{Corresponding author}
\affiliation{%
  \institution{City University of Hong Kong}
  \country{Hong Kong SAR}}
\email{chenma@cityu.edu.hk}


\renewcommand{\shortauthors}{Bowei He, et al.}

\begin{abstract}
Uplift modeling has been widely employed in online marketing by predicting the response difference between the treatment and control groups, so as to identify the sensitive individuals toward interventions like coupons or discounts. Compared with traditional \textit{conversion uplift modeling}, \textit{revenue uplift modeling} exhibits higher potential due to its direct connection with the corporate income. However, previous works can hardly handle the continuous long-tail response distribution in revenue uplift modeling. Moreover, they have neglected to optimize the uplift ranking among different individuals, which is actually the core of uplift modeling. To address such issues, in this paper, we first utilize the zero-inflated lognormal (ZILN) loss to regress the responses and customize the corresponding modeling network, which can be adapted to different existing uplift models. Then, we study the ranking-related uplift modeling error from the theoretical perspective and propose two tighter error bounds as the additional loss terms to the conventional response regression loss. Finally, we directly model the uplift ranking error for the entire population with a listwise uplift ranking loss. The experiment results on offline public and industrial datasets validate the effectiveness of our method for revenue uplift modeling. Furthermore, we conduct large-scale experiments on a prominent online fintech marketing platform, Tencent FiT, which further demonstrates the superiority of our method in real-world applications.
\end{abstract}


%

\maketitle
\vspace{-2mm}
\input{introduction.tex}

\input{related_work.tex}

\input{preliminaries.tex}

\input{methodology.tex}

\input{experiments.tex}

\input{conclusion.tex}

\begin{acks}
This work was supported by the Start-up Grant (No. 9610564), the Strategic Research Grant (No. 7005847) of the City University of Hong Kong, and the Early Career Scheme (No. CityU 21219323) of the University Grants Committee (UGC).
\end{acks}

\bibliographystyle{ACM-Reference-Format}
\bibliography{reference.bib}
\appendix
\input{appendix.tex}

\end{document}

%% file: introduction.tex
\section{Introduction}
\label{sec:intro}
Uplift modeling~\citep{gutierrez2017causal}, aiming to predict the expected difference between the treatment and control response, has been widely adopted to identify the individuals among a targeted population who can react positively to a particular intervention. Recently, this technique has been successfully deployed in many scenarios, like healthcare, finance transactions, and online marketing~\citep{shalit2017estimating, kane2014mining, zhou2023direct}. Generally, the uplift modeling in online marketing can be divided into two categories~\citep{DBLP:conf/amcis/GubelaL20}: \textit{conversion uplift modeling} and \textit{revenue uplift modeling}. The latter one is the focus of this work considering its higher application value and broader application scenarios.

Several methods have been proposed to perform uplift modeling to infer the causal effect of a specific treatment, such as price discount, repayment incentive, or coupon delivery. The most common one is the Randomized Controlled Trial (RCT)~\citep{sibbald1998understanding, deaton2018understanding} like the A/B test for marketing, where each individual is randomly assigned to the treatment group or the control group, thereby being independent of other covariate features. However, RCT experiments are often expensive, time-consuming, and even harmful to platforms in many applications. Even worse, RCT can only estimate the average uplift effect in the whole population, far from the individual uplift effect estimation, which is the pursuit of the current online marketing. Therefore, most recent works focus on learning an uplift model from the experimental data and then directly deploying it to estimate the uplift effect for studied individuals. Among them, the meta-learner methods~\citep{kunzel2019metalearners}, like S-Learner and T-Learner are the pioneer works seeking to estimate the uplift effect of personalized treatments on different individuals. However, they can be easily manipulated by the sample imbalance issue between the treatment and control group. Several tree-based methods~\citep{rzepakowski2010decision, davis2017using}, like Causal Forest, are proposed to address this issue by adapting the splitting criteria of conventional decision trees. Representation learning methods~\citep{shalit2017estimating, shi2019adapting} are another type of dominant approach to this issue with the development of deep neural networks especially in recent years. They mainly balance the representation distributions of individuals in the treatment group and control group.
\begin{figure}[ht]
    \centering
    \includegraphics[width=0.48\textwidth]{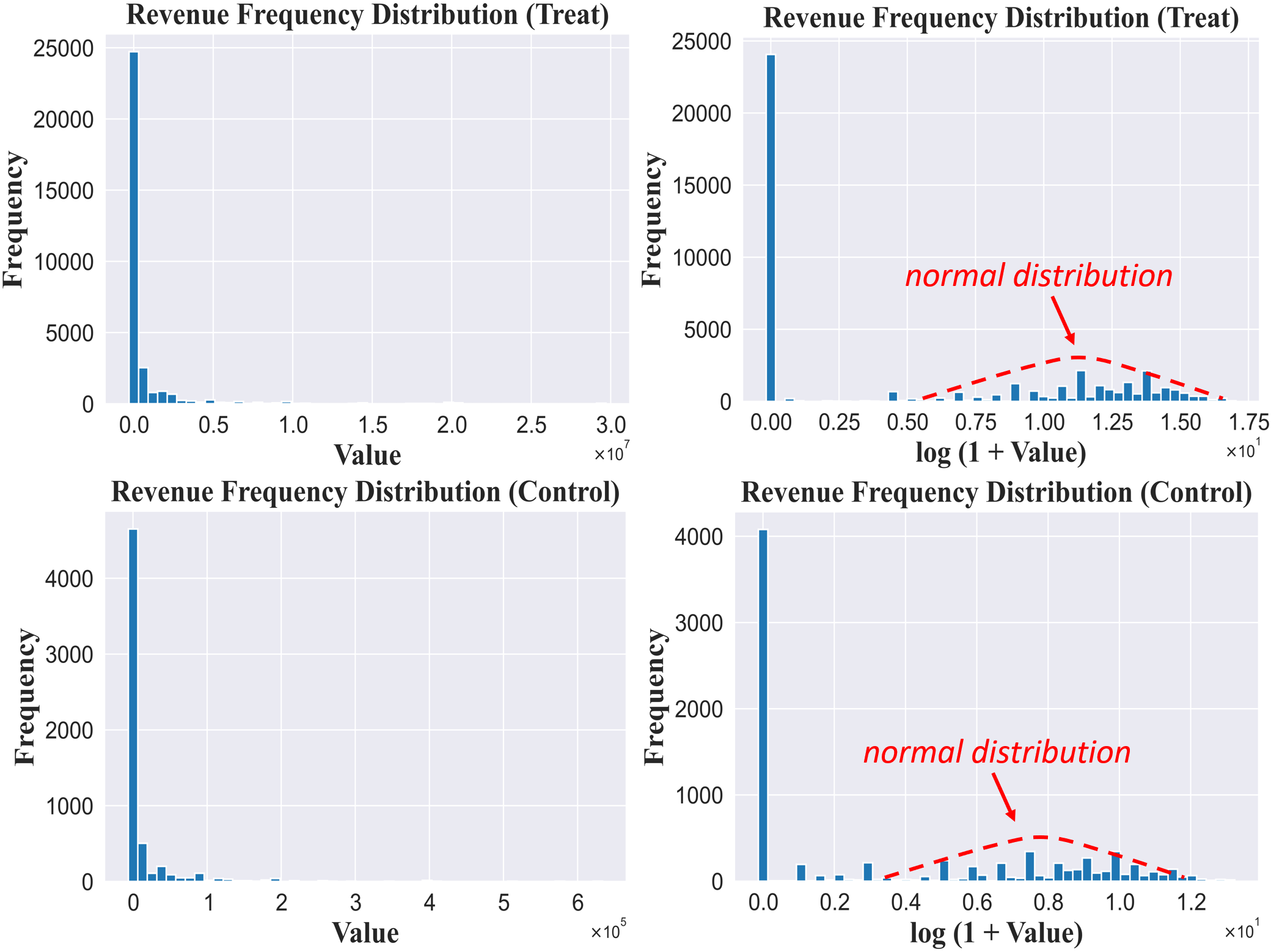}
    \caption{The histogram for the revenue frequency distribution of the A/B test in our deployment platform's online fintech application scenario. Left: raw revenue values; Right: revenue values after the logarithmic transformation.
    }
    \Description{The histogram for the revenue frequency distribution of the A/B test in our deployment platform's online fintech application scenario. Left: raw revenue values; Right: revenue values after the logarithmic transformation.}
    \label{fig: histogram}
    \vspace{-0.3cm}
\end{figure}

Though achieving acceptable performance in some idealized experiment environments or synthetic datasets, there are still some avenues to further improve the above methods. First, few of them focus on revenue uplift modeling, where the response is a \textit{continuous variable} rather than a simple binary one. This obviously increases the difficulty of precise uplift modeling. Meanwhile, this type of uplift modeling problem often has higher practical value though less explored, considering that practitioners hope to increase the return (continuous response) on marketing investments in many cases. Second, most of the existing methods~\citep{rzepakowski2010decision, shalit2017estimating, shi2019adapting, cui2024treatment} only work well on synthetic datasets, where the range of responses is limited, which is far from the real production environments, especially in finance scenarios. In such cases, the expected response, like the mutual fund sales revenue, can range from zero/several dollars to several million or even billion dollars. If directly adopting previous uplift methods to such tasks, some extreme data points can bring huge obstacles to model learning, thus the performance can hardly be satisfactory. Besides, the distribution of revenue response can be extremely imbalanced. In the above case, the sales revenue of most users hardly exceeds 10,000 USD, while only a very tiny proportion of users may spend over 500,000 USD. These two points can be concluded as the \textit{long-tail distribution} challenge, which is pretty common in online applications, like the one shown in Figure~\ref{fig: histogram}. Last but not least, almost all previous methods only care about the accuracy of the uplift effect prediction for different individuals, neglecting the ranking accuracy among them. That is, the individuals with higher true uplift values should obtain higher predicted uplift values compared with those lower-uplift value individuals, though the uplift value prediction is not so precise. In fact, the \textit{rankability of uplift models} is a more serious and meaningful problem in many real-world applications, because the ultimate purpose of utilizing the uplift models is to identify the individuals who are more susceptible to the treatment and then intervene with them differently according to the ranking of their uplift values. Moreover, this problem is more challenging in scenarios with \textit{long-tail} responses. Because in the learning phase, the model will mainly pay attention to individuals with high responses, and ignore the prediction accuracy of the individuals with lower responses. Thus, the uplift ranking confusion for these latter individuals can be exacerbated. 

To address the aforementioned challenges, we propose a \textbf{R}anking-\textbf{E}nhanced \textbf{R}evenue \textbf{U}plift \textbf{M}odeling (\textbf{RERUM}) framework. First, to overcome the continuous value and long-tail distribution challenges, we replace the conventional Mean-Squared-Error (MSE) loss with a ZILN loss for revenue response regression in treatment and control groups. Meanwhile, an accompanying regression network framework is proposed to adapt to ZILN loss. Second, we make a theoretical analysis of the uplift ranking error and propose more stringent error bounds as additional revenue response losses, which can help enhance the model's uplift ranking ability from the revenue response ranking perspective. Finally, we directly consider the uplift ranking among different individuals in the targeted population and propose a listwise ranking loss to explicitly optimize the uplift model's rankability.

To summarize, the main contributions of this work are as follows:
\begin{itemize}[leftmargin=*]
    \item To tackle the long-tail issue in revenue response regression, we propose the ZILN loss and its corresponding model framework.
    \item To mitigate the uplift modeling error, we perform a theoretical analysis and employ two tighter error bounds as response ranking losses to augment the above regression loss.
    \item To directly enhance the rankability of the uplift model over the whole population, we provide a listwise uplift ranking loss.
    \item To demonstrate the effectiveness of our method, we conduct extensive offline experiments on both public and industrial datasets. Additionally, large-scale online experiments on a fintech marketing platform with over 400 million users are also performed. 
\end{itemize}
\vspace{-4mm}

%% file: related_work.tex
\section{Related Works}
\label{sec: related works}
\textbf{Uplift Modeling}. The uplift modeling in online marketing typically comprise of two types of problems: \textit{conversion uplift modeling} and \textit{revenue uplift modeling}. In conversion uplift modeling~\citep{davis2017using, kunzel2019metalearners, kane2014mining, athey2015machine, betlei2021uplift}, the response label is a simple 0/1 binary variable indicating if the individual purchases the goods or takes a specific action, like clicking the advertisement. Most of previous research concentrates on this topic. Though achieving accepting performance in some scenarios, many of them~\citep{shalit2017estimating, shi2019adapting, li2022contrastive, wu2023stable} are dedicated to the uplift value prediction, neglecting the importance of uplift ranking among the population, which is originally the core point. Revenue uplift modeling~\citep{gubela2020response, DBLP:conf/amcis/GubelaL20, zhou2023direct}, however, differs from the traditional conversion uplift modeling on that its response label is a continuous variable with unlimited value range and irregular distribution, like the paying expense. This obviously introduces extra challenges for accurate uplift modeling. Unfortunately, few of works focus on the revenue uplift modeling problem, especially the rankability of revenue uplift models, though its higher application value and wider application scenarios in various online marketing platforms. In this paper, we dive into the rankability-enhanced revenue uplift modeling framework and explore its application in real-world scenarios.

\textbf{Learning to Rank}. Learning to rank (LTR) has raised growing attention in recent decades and have been deployed in many different applications. The \textit{pointwise} methods like Mcrank~\citep{li2007mcrank} are first proposed to solve this problem by transforming it to a regression or classification task on each single object. The models are trained to predict how relevant an object is for a given query. The core issue of this approach is that there exists the deviation between its optimization objective and the original ranking task. \textit{Pairwise} methods are an alternative solution which take object pairs into consideration and transform the ranking into a binary classification task. In other words, the models need to judge whether the order of each pair is wrong relative to the ground truth. The representative works in this line include: RankSVM~\citep{joachims2002optimizing}, Rankboost~\citep{freund2003efficient}, RankNet~\citep{burges2005learning}, LambdaRank~\citep{burges2006learning}, and LambdaMART~\citep{burges2010ranknet, hu2019unbiased}. \textit{Pairwise} methods have more promising results in practice because it comes closer to the nature of ranking than \textit{pointwise} ones. However, they also possess a significant drawback, i.e., ignoring the discrepancy among different pairs. To address this issue, \textit{listwise} methods like Adarank~\citep{xu2007adarank}, ListNet~\citep{cao2007learning}, ListMLE~\citep{xia2008listwise, lan2014position} have been proposed as a direct solution and consider all objects simultaneously. Nevertheless, most previous related literature is limited to the information retrieval tasks, like the document retrieval. Its application in causal effect estimation, especially the uplift modeling in online marketing is barely unexplored, though its great potential for improving the uplift model's ranking ability, which is the main focus of this work.

%% file: preliminaries.tex
\section{Preliminaries}
\label{sec:preliminary}
We follow the Neyman-Rubin potential outcome framework \citep{rubin2005causal} to formulate the revenue uplift modeling problem. 
In detail, we have the observed sample set $\mathcal{D}=\{(\mathbf{x}_i, t_i, y_i)\}^n_{i=1}$. For each individual $i \in \mathcal{I}$ ($|\mathcal{I}| = n$), $y_i\in \mathcal{Y}\subset \mathbb{R}$ is a continuous outcome (response), $\mathbf{x}_i \in \mathcal{X}\subset \mathbb{R}^d $ is a vector of covariates, and $t_i \in \mathcal{T} = \{0, 1\}$ denotes the treatment (with $t_i = 0$ as the control) intervened on individual $i$. $\mathit{Y}_i$, $\mathit{X}_i$, and $\mathit{T}_i$ are corresponding random variables, respectively. Furthermore, $\mathit{Y}$, $\mathit{X}$, and $\mathit{T}$ are the general random variables that indicate the response, covariate, and treatment, respectively, regardless of the specific individual. Thus, the uplift effect $\tau_i$ of the treatment on individual $i$ is defined as:
\begin{equation}
    \tau_i = y_i^{1} - y_i^{0},
\end{equation}
where $y_i^1$ and $y_i^0$ represent the potential responses under the corresponding treatment and control, respectively. Note that the whole population can be divided to two groups according to their received treatment: treatment group $\mathcal{D}^t = \{(\mathbf{x}_i, t_i, y_i)| (\mathbf{x}_i, t_i, y_i) \in \mathcal{D}, t_i = 1\}$, and control group $\mathcal{D}^c = \{(\mathbf{x}_i, t_i, y_i)| (\mathbf{x}_i, t_i, y_i) \in \mathcal{D}, t_i = 0\}$. $\mathbf{x}^{t/c}$ and $\mathit{X}^{t/c}$ represent the covariate vector instances of users in the treatment/control group and their corresponding random variable, respectively. 
However, due to the non-simultaneous observability of potential responses, the uplift effect in the Neyman-Rubin potential outcome framework can hardly be accessed directly. An empirical alternative is the Conditional Average Treatment Effect (CATE)~\citep{abrevaya2015estimating} which measures the response difference between the corresponding treatment and the control, conditioned on the observed covariates. In this paper, following previous works~\citep{betlei2021uplift, ma2022learning}, we estimate the CATE from the statistical perspective as the uplift effect, that is:
\begin{equation}
    \tau(\mathbf{x}_i) = \mathbb{E} [\mathbf{\mathit{Y}}_i | \mathbf{\mathit{X}}_i = \mathbf{x}_i, \mathbf{\mathit{T}}_i=1] - \mathbb{E} [\mathbf{\mathit{Y}}_i | \mathbf{\mathit{X}}_i = \mathbf{x}_i, \mathbf{\mathit{T}}_i=0].
\label{equ:tau-cate}
\end{equation}
It should be noted that when the uplift is estimated perfectly, the rankability of the model is also maximized.

%% file: methodology.tex
\section{Methodology}
In this section, we first introduce the uplift estimation model framework based on the previous related models and the newly designed zero-inflated lognormal loss for the revenue response regression. Second, we theoretically analyze the uplift modeling error from the ranking perspective and derive the tighter upper error bounds as the response ranking learning objectives. Then, motivated by the listwise ranking research, we directly model the uplift ranking among different individuals in the targeted population. Finally, we conclude the overall training objective for rankability-enhanced revenue uplift modeling.

\subsection{Uplift Estimation Model Framework}
\label{sec: uplift model framework}
According to the uplift effect formulation (Eq.~\ref{equ:tau-cate}) in Sec.~\ref{sec:preliminary}, conducting the accurate revenue response regression is undoubtedly crucial to the uplift modeling. However, previous uplift models~\citep{shalit2017estimating, shi2019adapting, li2022contrastive, wu2023stable} mainly rely on the MSE loss to learn response models which can hardly adapt to the continuous long-tail distribution in revenue uplift scenario, considering its sensitivity to outliers. To better achieve the goal of accurate response regression, we first study the response distribution. From Figure~\ref{fig: histogram}, we have two observations: 1) small values, especially zeros occupy the main frequency of the revenue response distribution, and 2) after the logarithmic transformation, except for such small (zero) values, the rest of the data roughly follows a normal distribution.

\begin{figure*}[ht]
    \centering
    \includegraphics[width=\textwidth]{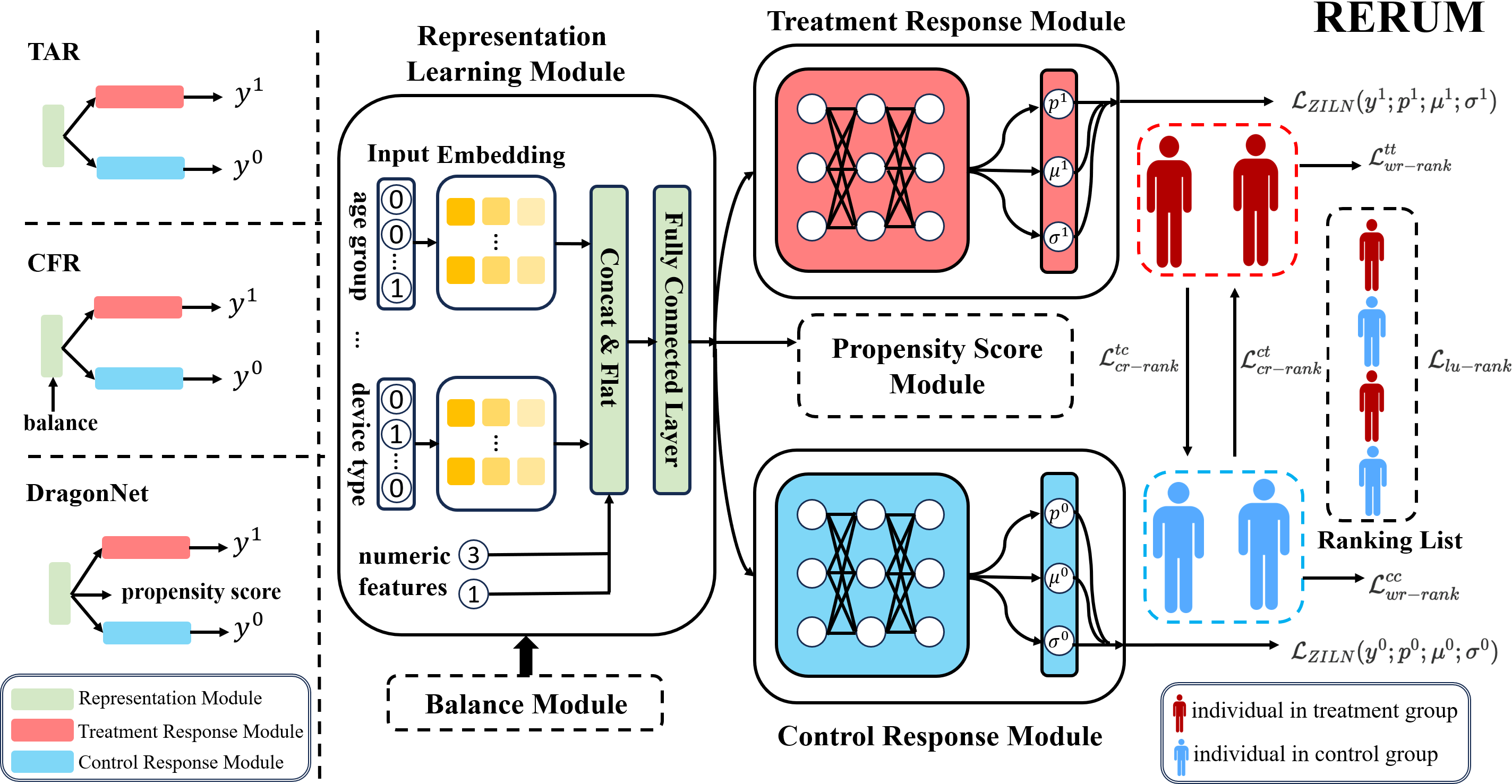}
    \caption{The overall framework of uplift estimation model. The modules in the dotted rectangular boxes are optional according to the specific base model selection. The original versions of base models take MSE as the response regression loss.
    }
    \Description{The overall framework of uplift estimation model. The modules in the dotted rectangular boxes are optional according to the specific base model selection. The original versions of base models take MSE as the response regression loss.}
    \label{fig: overall framework}
    \vspace{-0.3cm}
\end{figure*}

Motivated by the above observations, we refer to the related research in customer lifetime value research~\citep{wang2019deep} and propose the zero-inflated lognormal (ZILN) loss for response regression instead of the conventional MSE loss. In detail, based on the previous observation 1), the zero-inflation phenomenon, we first design a cross-entropy loss with the probability $p$ as the corresponding \textit{purchasing propensity} loss to help determine if the individual will spend money. Only the individuals that do spend money (payers) will engage into the computation of the following \textit{payer expense} loss. In this way, we can avoid the interference of such zero-value individuals in the positive revenue response regression. Furthermore, based on the above observation 2), we utilize the lognormal distribution with mean $\mu$ and standard deviation $\sigma$ to model the positive revenue response from such payers and correspondingly derive its negative log-likelihood as the \textit{payer expense} loss for non-zero response regression. In this lognormal loss, we conduct a logarithmic transformation to reduce the skewness of the response distribution. The mathematical form of our ZILN loss is as follows:
\begin{equation}
\begin{aligned}
    &\mathcal{L}_{ZILN}(y;p;\mu;\sigma) = \mathcal{L}_{purchasing \; propensity} + \mathcal{L}_{payer \; expense} \\
    &= \mathcal{L}_{CrossEntropy}(\mathbbm{1}(y>0); p) + \mathbbm{1}(y>0)\mathcal{L}_{Lognormal}(y;\mu;\sigma).
\end{aligned} 
\label{eq: reg loss}
\end{equation}
Furthermore, in this equation, we have
\begin{equation}
\begin{aligned}
     \mathcal{L}_{CrossEntropy}(\mathbbm{1}(y>0); p) &= - \mathbbm{1}(y=0)\text{log}(1-p) - \mathbbm{1}(y>0)\text{log}p, \\
     \mathcal{L}_{Lognormal}(y;\mu;\sigma) &= \text{log}(y\sigma\sqrt{2\pi}) + \frac{(\text{log}y-\mu)^2)}{2\sigma^2}.
\end{aligned}  
\label{eq:ziln}
\end{equation}

It should be mentioned that the ZILN loss can not only facilitate the agreement between the predicted response and the true response, but also help discriminate the ranking among the responses~\citep{wang2019deep}, which is beneficial to the response ranking learning module in the later Sec.~\ref{sec:rrl}.

To match the realization of the above ZILN loss, we customize the uplift estimation model framework based on the previous related uplift models shown on the left side of Figure~\ref{fig: overall framework}. As shown in the right side of  Figure~\ref{fig: overall framework}, our newly designed framework includes the representation learning module, treatment response module, control response module, etc. For a given individual, the treatment/control response module will take the corresponding representation vector from the representation learning module to predict $p^1, \mu^1, \sigma^1$ or $p^0, \mu^0, \sigma^0$, respectively, which will be further utilized to obtain regression loss via Eq.~\ref{eq: reg loss}. Note our proposed framework is agnostic to the specific uplift model choice; any previously mentioned base models in Sec.~\ref{sec: related works} can be integrated with our framework. In our following implementations, we mainly adopt the deep learning-based models as the workhorse due to their strong ability to capture the complex and nonlinear relations between covariates and responses, as well as the competitive performance proved by previous works.

\subsection{Uplift Modeling Error Analysis}
\label{sec: error analysis}
To restore the correct ranking among the $\tau_i$ for different individuals, an intuitive approach is to ensure that the predicted uplift distance $\hat{\tau}_i - \hat{\tau}_j$ between each possible individual pair $(i, j)$ is as close as possible to their true uplift distance $\tau_i - \tau_j$. Therefore, in the following part, we focus on the uplift distance prediction error $|(\hat{\tau}_i - \hat{\tau}_j) - (\tau_i - \tau_j)|$ and theoretically prove that conventional MSE loss for response regression is a loose error bound which can hardly capture the true uplift distance between individual pairs.
\begin{proof}
Take randomly selected individuals $i$ and $j$ as an example. Firstly, according to the definition of the uplift effect, we have the following equations:
\begin{equation}
\begin{aligned}
    \hat{\tau}_i - \hat{\tau}_j &= (\hat{y}^1_i - \hat{y}^0_i) - (\hat{y}^1_j - \hat{y}^0_j)  = (\hat{y}^1_i - \hat{y}^1_j) - (\hat{y}^0_i - \hat{y}^0_j), \\
    \tau_i - \tau_j &= (y^1_i - y^0_i) - (y^1_j - y^0_j)  = (y^1_i - y^1_j) - (y^0_i - y^0_j).
\end{aligned}  
\label{eq:error1}
\end{equation}
Thus, we can obtain the discrepancy between the predicted pairwise uplift distance and the true pairwise uplift distance for individual $i$ and $j$ as follows:
\begin{equation}
\begin{aligned}
    &\quad |(\hat{\tau}_i - \hat{\tau}_j) - (\tau_i - \tau_j)|  \\
    &= |((\hat{y}^1_i - \hat{y}^1_j) - (\hat{y}^0_i - \hat{y}^0_j)) - ((y^1_i - y^1_j) - (y^0_i - y^0_j))| \\
    &= |((\hat{y}^1_i - \hat{y}^1_j) - (y^1_i - y^1_j)) - ((\hat{y}^0_i - \hat{y}^0_j) - (y^0_i - y^0_j))| \\
    &\leq |(\hat{y}^1_i - \hat{y}^1_j) - (y^1_i - y^1_j)| + |(\hat{y}^0_i - \hat{y}^0_j) - (y^0_i - y^0_j)|. 
\end{aligned}
\label{eq:error2}
\end{equation}
Based on the above derivation, we have:
\begin{equation}
\begin{aligned}
    &\quad |(\hat{\tau}_i - \hat{\tau}_j) - (\tau_i - \tau_j)|  \\
    &\leq |(\hat{y}^1_i - \hat{y}^1_j) - (y^1_i - y^1_j)| + |(\hat{y}^0_i - \hat{y}^0_j) - (y^0_i - y^0_j)| \\
    &= |(\hat{y}^1_i - y^1_i) - (\hat{y}^1_j - y^1_j)| + |(\hat{y}^0_i - y^0_i) - (\hat{y}^0_j - y^0_j)| \\
    &\leq |\hat{y}^1_i - y^1_i| + |\hat{y}^1_j - y^1_j| + |\hat{y}^0_i - y^0_i| + |\hat{y}^0_j - y^0_j|.
\end{aligned}
\label{eq:error3}
\end{equation}
Meanwhile, another equation variant similar to Eq.~\ref{eq:error2} is as follows:
\begin{equation}
\begin{aligned}
    &\quad |(\hat{\tau}_i - \hat{\tau}_j) - (\tau_i - \tau_j)|  \\
    &= |((\hat{y}^1_i - \hat{y}^1_j) - (\hat{y}^0_i - \hat{y}^0_j)) - ((y^1_i - y^1_j) - (y^0_i - y^0_j))| \\
    &= |((\hat{y}^1_i - y^0_j) - (y^1_i - \hat{y}^0_j)) - ((\hat{y}^0_i - y^1_j ) - (y^0_i - \hat{y}^1_j))| \\
    &\leq |(\hat{y}^1_i - y^0_j) - (y^1_i - \hat{y}^0_j)| + |(\hat{y}^0_i - y^1_j ) - (y^0_i - \hat{y}^1_j)|.
\end{aligned}
\label{eq:error4}
\end{equation}
Following Eq.~\ref{eq:error4}, we can derive:
\begin{equation}
\begin{aligned}
    &\quad |(\hat{\tau}_i - \hat{\tau}_j) - (\tau_i - \tau_j)|  \\
    &\leq |(\hat{y}^1_i - y^0_j) - (y^1_i - \hat{y}^0_j)| + |(\hat{y}^0_i - y^1_j ) - (y^0_i - \hat{y}^1_j)|\\
    &= |(\hat{y}^1_i - y^1_i) + (\hat{y}^0_j - y^0_j)| + |(\hat{y}^0_i - y^0_i) + (\hat{y}^1_j - y^1_j)| \\
    &\leq |\hat{y}^1_i - y^1_i| + |\hat{y}^0_j - y^0_j| + | \hat{y}^0_i - y^0_i| + |\hat{y}^1_j - y^1_j|.
\end{aligned}
\label{eq:error5}
\end{equation}
\vspace{-0.3cm}
\end{proof}

From the Eq.~\ref{eq:error3} and~\ref{eq:error5}, we can find their last step $|\hat{y}^1_i - y^1_i| + |\hat{y}^1_j - y^1_j| + |\hat{y}^0_i - y^0_i| + |\hat{y}^0_j - y^0_j|$ are just equivalent to the optimization objective of response MSE loss. This means that conventional MSE loss for response modeling is actually a looser upper error bound than $|(\hat{y}^1_i - \hat{y}^1_j) - (y^1_i - y^1_j)| + |(\hat{y}^0_i - \hat{y}^0_j) - (y^0_i - y^0_j)| $ and $|(\hat{y}^1_i - y^0_j) - (y^1_i - \hat{y}^0_j)| + |(\hat{y}^0_i - y^1_j ) - (y^0_i - \hat{y}^1_j)|$ from the perspective of uplift ranking, while the latter two are the focus of our following response ranking learning.

\subsection{Response Ranking Learning}
\label{sec:rrl}
From the above analysis in Sec.~\ref{sec: error analysis}, we obtain two tighter uplift modeling error bounds from the ranking perspective. Thus, we can transform them into the within-group response ranking loss and cross-group ranking loss between the predicted and the real responses, respectively, to enhance the uplift model's rankability. 
\subsubsection{Within-group Response Ranking Loss}
We first design the with-group response loss as follows according to Eq.~\ref{eq:error2}:
\begin{equation}
\begin{aligned}
\mathcal{L}^{tt/cc}_{wr-rank} (i, j) = \left \{
\begin{aligned}
     &0,  \quad \quad \quad if (\hat{y}^{1/0}_i - \hat{y}^{1/0}_j) \cdot (y^{1/0}_i - y^{1/0}_j) \geq 0 \\
     &((\hat{y}^{1/0}_i - \hat{y}^{1/0}_j) - (y^{1/0}_i - y^{1/0}_j))^2,  otherwise. 
\end{aligned}  
\right.
\end{aligned}   
\label{eq:wr-ranking}
\end{equation}
In Eq.~\ref{eq:wr-ranking}, for the individual pair $(i, j)$,  if the relative order between predicted responses differs from that between true responses, we punish the uplift model with the above alignment loss. Adding the within-group response ranking loss in the treatment group $\mathcal{L}^{tt}_{wr-rank}$ to the within-group response ranking loss in the control group $\mathcal{L}^{cc}_{wr-rank}$, we can obtain the overall $\mathcal{L}_{wr-rank}$:
\begin{equation}
\begin{aligned}
\mathcal{L}_{wr-rank} = \underset{i,j\leq|\mathcal{D}^t|}{\sum} \mathcal{L}^{tt}_{wr-rank} (i, j) + \underset{i,j\leq|\mathcal{D}^c|}{\sum} \mathcal{L}^{cc}_{wr-rank} (i, j).
\end{aligned}    
\end{equation}
\subsubsection{Cross-group Response Ranking Loss}
Then, we further devise the cross-group response ranking loss as follows according to Eq.~\ref{eq:error4}:
\begin{equation}
\begin{aligned}
\mathcal{L}^{tc/ct}_{cr-rank} (i, j) = \left \{
\begin{aligned}
     &0,  \quad \quad \quad if (\hat{y}^{1/0}_i - y^{0/1}_j) * (y^{1/0}_i - \hat{y}^{0/1}_j) \geq 0 \\
     &((\hat{y}^{1/0}_i - y^{0/1}_j) - (y^{1/0}_i - \hat{y}^{0/1}_j))^2,  otherwise. 
\end{aligned}  
\right.
\end{aligned}   
\label{eq:cr-ranking}
\end{equation}
Similarly, we punish the uplift model when the predicted responses of individual pair $(i, j)$ cannot meet the cross-group response-related order criteria. Then, adding the treatment-control response ranking loss $\mathcal{L}^{tc}_{cr-rank}$ to the control-treatment response ranking loss $\mathcal{L}^{ct}_{cr-rank}$, we obtain:
\begin{equation}
\begin{aligned}
\mathcal{L}_{cr-rank} = \underset{\substack{i\leq|\mathcal{D}^t|, \\j\leq|\mathcal{D}^c|}}{\sum} \mathcal{L}^{tc}_{cr-rank} (i, j) + \underset{\substack{i\leq|\mathcal{D}^c|,\\ j\leq|\mathcal{D}^t|}}{\sum} \mathcal{L}^{ct}_{cr-rank} (i, j).
\end{aligned}    
\end{equation}
Combining the $\mathcal{L}_{wr-rank}$ and $\mathcal{L}_{cr-rank}$, the final response ranking loss $\mathcal{L}_{r-rank}$ is acquired:
\begin{equation}
\begin{aligned}
\mathcal{L}_{r-rank} = \mathcal{L}_{wr-rank} + \mathcal{L}_{cr-rank}.
\end{aligned}    
\end{equation}
Note that in the practical implementation of above $\mathcal{L}_{wr-rank}$ and $\mathcal{L}_{cr-rank}$, we randomly sample a certain number of individuals in each group rather than traversing all individuals to accelerate the computation process.

\subsection{Uplift Ranking Learning}
\label{sec:url}
In this section, we directly dive into the uplift ranking problem and explicitly optimize the uplift model's rankability. It can be easily noted that this is inherently a learning-to-rank problem. The previous solutions mainly include three types: pointwise, pairwise, and listwise. The former two kinds consider the ranking problem from the perspective of a single individual or an individual pair, thus can hardly take all individuals into consideration at a glance. Besides, the pairwise approaches are mainly classification-based and thus have great difficulty distinguishing the inner-pair uplift distance among different pairs. Therefore, they are more suitable for binary responses, i.e., conversion uplift problems. If they are deployed in the revenue uplift scenario where the response is continuous, the performance can hardly be satisfying. The listwise approach, however, can directly minimize the overall ranking error of all individuals in the population and discern the difference in continuous revenue uplift values corresponding to different individuals.

Therefore, we follow the listwise ranking procedure and first model the uplift ranking process as a random process, which means each individual has the probability to be ranked in any position of the whole population list. Here, the ranking probability on the higher position for an individual should be positively related to her uplift score. To tackle this problem, we start from the probability that an individual is ranked at the top-one position of the whole list and propose the loss function to measure the uplift model ranking error. Then, we derive such loss function to an optimizable form with the given data.

\subsubsection{Uplift Top One Probability}
Here, referring to~\citep{cao2007learning}, we first define the probability that a random individual with the covariate $\mathit{X}$ to be ranked at the top one position:
\begin{equation}
\begin{aligned}
 p_{\tau}(\mathit{X}) &= \frac{e^{s(\tau(\mathit{X}))}}{\sum_{\mathit{X}} e^{s(\tau(\mathit{X}))}},
\end{aligned}
\label{eq:topone}
\end{equation}
where $s(\cdot)$ is the ranking score function based on the uplift score for $\mathit{X}$. And it should possess the monotonically increasing property.

\subsubsection{Listwise Uplift Ranking Loss}
With the above definition, given the list of true uplift score $\tau$ and the list of predicted uplift score $\hat{\tau}$ by the model, we can measure the distance between them as the overall ranking error. Here, we utilize the Cross-Entropy as the measurement and define the listwise uplift ranking loss as follows:
\begin{equation}
\begin{aligned}
 \mathcal{L}_{lu-rank} &= - \underset{\mathit{X}}{\mathbb{E}} \: p_{\tau}(\mathit{X}) \: ln (p_{\hat{\tau}}(\mathit{X})).
\end{aligned}
\label{eq:lu-rank}
\end{equation}
To transform Eq.~\ref{eq:lu-rank} to an optimizable form, we provide the following derivation process:
\begin{proof}
\begin{equation}
\begin{aligned}
 \mathcal{L}_{lu-rank} &= - \underset{\mathit{X}}{\mathbb{E}} \: p_{\tau}(\mathit{X}) \: ln (p_{\hat{\tau}}(\mathit{X}))\\
 & = - \underset{\mathit{X}}{\mathbb{E}} \: \frac{e^{s(\tau(\mathit{X}))}}{\sum_{\mathit{X}} e^{s(\tau(\mathit{X}))}} \: ln (\frac{e^{\hat{s}(\hat{\tau}(\mathit{X}))}}{\sum_{\mathit{X}} e^{\hat{s}(\hat{\tau}(\mathit{X}))}}).
\end{aligned}
\end{equation}
Here, we take $s(\tau(\mathit{X})) = ln (\tau(x)), \hat{s}(\hat{\tau}(\mathit{X})) = \hat{\tau}(x)$. Thus, we have:
\begin{equation}
\begin{aligned}
 \mathcal{L}_{lu-rank} &= - \underset{\mathit{X}}{\mathbb{E}} \: \frac{\tau(\mathit{X})}{\sum_{\mathit{X}}\tau(\mathit{X})} \: ln(\frac{e^{\hat{\tau}(\mathit{X})}}{\sum_{\mathit{X}}e^{\hat{\tau}(\mathit{X})}}).
\end{aligned}
\end{equation}
Considering that $\sum_{\mathit{X}}\tau(\mathit{X})$ is an unknown constant for each instance of variable $\mathit{X}$, we remove it from the equation to facilitate the implementation. 
\begin{equation}
\begin{aligned}
 \mathcal{L}_{lu-rank} &= - \underset{\mathit{X}}{\mathbb{E}} \: \tau(\mathit{X}) \: ln(\frac{e^{\hat{\tau}(\mathit{X})}}{\sum_{\mathit{X}}e^{\hat{\tau}(\mathit{X})}})\\
 & = - \underset{\mathit{X}}{\mathbb{E}} \: (\mathit{Y}^1(\mathit{X}) - \mathit{Y}^0(\mathit{X})) \: ln(\frac{e^{\hat{\tau}(\mathit{X})}}{\sum_{\mathit{X}}e^{\hat{\tau}(\mathit{X})}}) \\
  & = - (\underset{\mathit{X}}{\mathbb{E}} \: \mathit{Y}^1(\mathit{X}) \: ln(\frac{e^{\hat{\tau}(\mathit{X})}}{\sum_{\mathit{X}}e^{\hat{\tau}(\mathit{X})}})  - \underset{\mathit{X}}{\mathbb{E}} \: \mathit{Y}^0(\mathit{X}) \: ln(\frac{e^{\hat{\tau}(\mathit{X})}}{\sum_{\mathit{X}}e^{\hat{\tau}(\mathit{X})}})).  \\
\end{aligned}
\end{equation}
Under the setting of RCT experiments, the assumption that $\mathit{X}$, $\mathit{X}^t$, and $\mathit{X}^c$ follow the same distribution holds. Thus, we can obtain:
\begin{equation}
\begin{aligned}
 \mathcal{L}_{lu-rank}  &= - (\underset{\mathit{X^t}}{\mathbb{E}} \: \mathit{Y}^1(\mathit{X^t}) \: ln(\frac{e^{\hat{\tau}(\mathit{X^t})}}{\sum_{\mathit{X}}e^{\hat{\tau}(\mathit{X})}})  - \underset{\mathit{X^c}}{\mathbb{E}} \: \mathit{Y}^0(\mathit{X^c}) \: ln(\frac{e^{\hat{\tau}(\mathit{X^c})}}{\sum_{\mathit{X}}e^{\hat{\tau}(\mathit{X})}}))
 \\
  &= -(\frac{1}{|\mathcal{D}^t|} \underset{\mathbf{x}^t \in \mathcal{D}^t}{\sum} y^1(\mathit{\mathbf{x}^t}) \: ln(\frac{e^{\hat{\tau}(\mathbf{x}^t)}}{\underset{\mathbf{x} \in \mathcal{D}}{\sum} e^{\hat{\tau}(\mathbf{x})}}) \\
  & \quad - \frac{1}{|\mathcal{D}^c|} \underset{\mathbf{x}^c \in \mathcal{D}^c}{\sum} y^0(\mathbf{x}^c) \: ln(\frac{e^{\hat{\tau}(\mathbf{x}^c)}}{\underset{\mathbf{x} \in \mathcal{D}}{\sum} e^{\hat{\tau}(\mathbf{x})}})).
\end{aligned}
\label{eq:lu-rank train}
\end{equation}
\end{proof}
\vspace{-0.3cm}

\subsection{Overall Training Objective}
After obtaining $\mathcal{L}_{ZILN}$ in Sec.~\ref{sec: uplift model framework}, $\mathcal{L}_{r-rank}$ in Sec.~\ref{sec:rrl}, and $\mathcal{L}_{lu-rank}$ in Sec.~\ref{sec:url} sequentially, we add them together and train the parameters $\bm{\theta}$ of uplift model $f$ with the gradient descent algorithm:
\begin{equation}
\begin{aligned}
 \mathcal{L}_{overall} &= \mathcal{L}_{ZILN} + \mathcal{L}_{r-rank} + \mathcal{L}_{lu-rank} + \lambda \Vert \bm{\theta} \Vert^2_2,
\end{aligned}
\label{eq:overall}
\end{equation}
where $\lambda$ is the coefficient of the L2 regularization term. To better illustrate the process flow of our proposed RERUM framework, we provide the pseudocode in Algorithm~\ref{alg:algorithm}.
\begin{algorithm}
 \caption{RERUM}
 \label{alg:algorithm}
 \KwIn{The observed sample set $\mathcal{D}$ consisting of treatment group $\mathcal{D}^t$ and control group $\mathcal{D}^c$, sample number $S$}
 \KwOut{Feature embeddings $\mathbf{e}_f$, representation learning module parameters $\mathbf{\theta}_r$, treatment/control response module parameters $\mathbf{\theta}_t$, $\mathbf{\theta}_c$.}
 \textbf{Process}:\;
Initialize embeddings and parameters $\mathbf{e}_f, \mathbf{\theta}_r, \mathbf{\theta}_t, \mathbf{\theta}_c$\;
\While{Not Converged}{
       \For{batch $\mathcal{B}^t$ in $\mathcal{D}^t$, batch $\mathcal{B}^c$ in $\mathcal{D}^c$}{
        Obtain predicted $(p^1, \mu^1, \sigma^1)$ and $(p^0, \mu^0, \sigma^0)$ for each sample in $\mathcal{B}^t$ and $\mathcal{B}^c$, respectively\;
        Compute the ZILN regression loss $\mathcal{L}_{ZILN}^t$ and $\mathcal{L}_{ZILN}^c$ for $\mathcal{B}^t$ and $\mathcal{B}^c$ with Eqn.~\ref{eq: reg loss} \;
        Sample $S$ individuals from $\mathcal{B}^t$ and $\mathcal{B}^c$ separately and compute $\mathcal{L}^{tt}_{wr-rank}$ and $\mathcal{L}^{cc}_{wr-rank}$ with Eqn.~\ref{eq:wr-ranking} \;
        Sample $S$ individuals from $\mathcal{B}^t$ and $\mathcal{B}^c$ separately and compute $\mathcal{L}^{tc}_{cr-rank}$ and $\mathcal{L}^{ct}_{cr-rank}$ with Eqn.~\ref{eq:cr-ranking} \;
        Utilize $\mathcal{B}^t$ and $\mathcal{B}^c$ to compute the in-batch listwise uplift ranking loss $\mathcal{L}_{lu-rank}$ with Eqn.~\ref{eq:lu-rank train} \;
        Add above loss terms together to obtain $\mathcal{L}_{overall}$ with Eqn.~\ref{eq:overall} \;
        Update parameters $\mathbf{e}_f, \mathbf{\theta}_r, \mathbf{\theta}_t, \mathbf{\theta}_c$ with end-to-end gradient descent on $\mathcal{L}_{overall}$ \;
        }}
\end{algorithm}
\vspace{-3mm}

%% file: experiments.tex
\section{Experiments}
\label{sec:experiments}
In this section, we conduct experiments on three offline datasets and an online fintech marketing platform to show the effectiveness of our method. We mainly focus on the following questions:
\begin{itemize}[leftmargin=*]
\item \textbf{RQ1:} Can our method outperform different baselines on various ranking-related uplift modeling metrics?

\item \textbf{RQ2:} How does each proposed module contribute to the overall revenue uplift modeling performance?

\item \textbf{RQ3:} How do the hyper-parameters influence the performance of our method in different datasets?

\item \textbf{RQ4:} How do our proposed modules influence the response regression performance?


\item \textbf{RQ5:} How does our method perform in the online deployment scenario compared with other online service strategies?

\end{itemize}

\begin{table*}[!t]
\caption{Overall performance of different methods on different uplift-ranking related metrics in various datasets. The best method and best baseline on each metric are marked as bold and underlined, respectively. $\uparrow$ refers to the improvement over the best-performing baseline. $*$ indicates the improvements over baselines are statistically significant ($t$-test, $p$-value $\leq 0.05$). For ease of illustration, the LIFT@30 for Product dataset has been divided by 10,000.}
\resizebox{0.99\textwidth}{!}{
\begin{tabular}{l|cccc|cccc|cccc}
\toprule
\multicolumn{1}{c|}{\multirow{2}{*}{Methods}} & \multicolumn{4}{c|}{Hillstrom-Men}                                                          & \multicolumn{4}{c|}{Hillstrom-Women}                                                        & \multicolumn{4}{c}{Product}                                                                \\ \cline{2-13} 
\multicolumn{1}{c|}{}                         & \multicolumn{1}{c|}{AUUC} & \multicolumn{1}{c|}{AUQC} & \multicolumn{1}{c|}{KRCC} & LIFT@30 & \multicolumn{1}{c|}{AUUC} & \multicolumn{1}{c|}{AUQC} & \multicolumn{1}{c|}{KRCC} & LIFT@30 & \multicolumn{1}{c|}{AUUC} & \multicolumn{1}{c|}{AUQC} & \multicolumn{1}{c|}{KRCC} & LIFT@30 \\ \midrule
Causal Forest                                      & \multicolumn{1}{c|}{0.4492}     & \multicolumn{1}{c|}{0.4555}     & \multicolumn{1}{c|}{0.0120}     &  0.6376       & \multicolumn{1}{c|}{0.5653}     & \multicolumn{1}{c|}{0.5710}     & \multicolumn{1}{c|}{0.0776}     &    0.3260     & \multicolumn{1}{c|}{0.6625}     & \multicolumn{1}{c|}{0.6809}     & \multicolumn{1}{c|}{0.3265}     &  0.9252       \\ 
S-Learner                                       & \multicolumn{1}{c|}{0.5301}     & \multicolumn{1}{c|}{0.5297}     & \multicolumn{1}{c|}{0.0485}     &   0.9123      & \multicolumn{1}{c|}{0.5537}     & \multicolumn{1}{c|}{0.5554}     & \multicolumn{1}{c|}{0.0654}     &  0.4127       & \multicolumn{1}{c|}{0.6107}     & \multicolumn{1}{c|}{0.6088}     & \multicolumn{1}{c|}{0.2017}     &   0.8451      \\ 
T-Learner                                  & \multicolumn{1}{c|}{0.5565}     & \multicolumn{1}{c|}{0.5594}     & \multicolumn{1}{c|}{0.0327}     &     1.0417    & \multicolumn{1}{c|}{0.5733}     & \multicolumn{1}{c|}{0.5735}     & \multicolumn{1}{c|}{0.1018}     &   0.6564      & \multicolumn{1}{c|}{0.6212}     & \multicolumn{1}{c|}{0.6374}     & \multicolumn{1}{c|}{0.2238}     &   0.8941      \\ 
TR                                             & \multicolumn{1}{c|}{0.5484}     & \multicolumn{1}{c|}{0.5493}     & \multicolumn{1}{c|}{0.0309}     &  1.0388       & \multicolumn{1}{c|}{0.5801}     & \multicolumn{1}{c|}{0.5814}     & \multicolumn{1}{c|}{0.1317}     &    0.4315     & \multicolumn{1}{c|}{0.6683}     & \multicolumn{1}{c|}{0.6983}     & \multicolumn{1}{c|}{0.3337}     &  \underline{0.9429}       \\ 
TAR                                         & \multicolumn{1}{c|}{0.5652}     & \multicolumn{1}{c|}{0.5659}     & \multicolumn{1}{c|}{0.0780}     &   0.7974      & \multicolumn{1}{c|}{0.5739}     & \multicolumn{1}{c|}{0.5770}     & \multicolumn{1}{c|}{0.0586}     &  0.6366       & \multicolumn{1}{c|}{0.6493}     & \multicolumn{1}{c|}{0.6737}     & \multicolumn{1}{c|}{0.2986}     & 0.8918        \\ 
CFR$_{wass}$                                            & \multicolumn{1}{c|}{0.5661}     & \multicolumn{1}{c|}{0.5676}     & \multicolumn{1}{c|}{0.0788}     &    0.8721     & \multicolumn{1}{c|}{0.5754}     & \multicolumn{1}{c|}{0.5762}     & \multicolumn{1}{c|}{0.0606}     &   0.6472      & \multicolumn{1}{c|}{0.6957}     & \multicolumn{1}{c|}{0.7018}     & \multicolumn{1}{c|}{0.2998}     &  0.8741       \\ 
CFR$_{mmd}$                                            & \multicolumn{1}{c|}{0.5760}     & \multicolumn{1}{c|}{0.5762}     & \multicolumn{1}{c|}{0.0747}     &  0.7351       & \multicolumn{1}{c|}{0.5836}     & \multicolumn{1}{c|}{0.5814}     & \multicolumn{1}{c|}{0.0788}     &   0.6206      & \multicolumn{1}{c|}{0.6933}     & \multicolumn{1}{c|}{0.7025}     & \multicolumn{1}{c|}{\underline{0.3430}}     &  0.8929       \\ 
StableCFR                                      & \multicolumn{1}{c|}{0.5772}     & \multicolumn{1}{c|}{0.5783}     & \multicolumn{1}{c|}{0.0786}     &    0.7524     & \multicolumn{1}{c|}{0.5725}     & \multicolumn{1}{c|}{0.5687}     & \multicolumn{1}{c|}{0.0965}     &  0.6057       & \multicolumn{1}{c|}{0.6948}     & \multicolumn{1}{c|}{\underline{0.7067}}     & \multicolumn{1}{c|}{0.3162}     &   0.8547      \\ 
CITE                                           & \multicolumn{1}{c|}{0.5812}     & \multicolumn{1}{c|}{0.5793}     & \multicolumn{1}{c|}{0.0827}     &  0.7928       & \multicolumn{1}{c|}{0.5785}     & \multicolumn{1}{c|}{0.5717}     & \multicolumn{1}{c|}{0.1147}     &    0.6742     & \multicolumn{1}{c|}{\underline{0.6976}}     & \multicolumn{1}{c|}{0.6987}     & \multicolumn{1}{c|}{0.3348}     &  0.8738       \\ 
DragonNet                                      & \multicolumn{1}{c|}{\underline{0.6028}}     & \multicolumn{1}{c|}{\underline{0.6042}}     & \multicolumn{1}{c|}{\underline{0.0897}}     &  \underline{1.3526}       & \multicolumn{1}{c|}{\underline{0.5858}}     & \multicolumn{1}{c|}{\underline{0.5836}}     & \multicolumn{1}{c|}{\underline{0.1368}}     &  \underline{0.7123}       & \multicolumn{1}{c|}{0.6347}     & \multicolumn{1}{c|}{0.6568}     & \multicolumn{1}{c|}{0.3253}     &  0.8148       \\ \midrule
RERUM (TAR)                                       & \multicolumn{1}{c|}{\begin{tabular}[c]{@{}c@{}} 0.6299*\\ $\uparrow$ 4.50\%\end{tabular}}     & \multicolumn{1}{c|}{\begin{tabular}[c]{@{}c@{}} 0.6338*\\ $\uparrow$ 4.90\% \end{tabular}}     & \multicolumn{1}{c|}{\begin{tabular}[c]{@{}c@{}}0.1617*\\ $\uparrow$ 80.27\%\end{tabular}}     &   \begin{tabular}[c]{@{}c@{}} 1.3477*\\ $\downarrow$ 0.36\% \end{tabular}      & \multicolumn{1}{c|}{\begin{tabular}[c]{@{}c@{}} 0.6360*\\ $\uparrow$ 8.57\% \end{tabular}}     & \multicolumn{1}{c|}{\begin{tabular}[c]{@{}c@{}} 0.6334*\\ $\uparrow$ 8.53\% \end{tabular}}     & \multicolumn{1}{c|}{\begin{tabular}[c]{@{}c@{}} 0.1806*\\ $\uparrow$ 32.02\% \end{tabular}}     &   \begin{tabular}[c]{@{}c@{}} 0.8755*\\ $\uparrow$ 22.91\% \end{tabular}      & \multicolumn{1}{c|}{\begin{tabular}[c]{@{}c@{}}0.6863*\\ $\downarrow$ 1.62\%\end{tabular}}     & \multicolumn{1}{c|}{\begin{tabular}[c]{@{}c@{}}0.7086*\\ $\uparrow$ 0.27\%\end{tabular}}     & \multicolumn{1}{c|}{\begin{tabular}[c]{@{}c@{}}0.3269*\\ $\downarrow$ 4.69\%\end{tabular}}     &    \begin{tabular}[c]{@{}c@{}}1.0247*\\ $\uparrow$ 8.68\%\end{tabular}     \\ \midrule
RERUM (CFR$_{wass}$)                                       & \multicolumn{1}{c|}{\begin{tabular}[c]{@{}c@{}} 0.6474*\\ $\uparrow$ 7.40\%\end{tabular}}     & \multicolumn{1}{c|}{\begin{tabular}[c]{@{}c@{}} 0.6497*\\ $\uparrow$ 7.53\%\end{tabular}}     & \multicolumn{1}{c|}{\begin{tabular}[c]{@{}c@{}}0.1382*\\ $\uparrow$ 54.07\% \end{tabular}}     &   \begin{tabular}[c]{@{}c@{}} 1.4602*\\ $\uparrow$ 7.96\% \end{tabular}      & \multicolumn{1}{c|}{\begin{tabular}[c]{@{}c@{}} 0.6568*\\ $\uparrow$ 12.12\%\end{tabular}}     & \multicolumn{1}{c|}{\begin{tabular}[c]{@{}c@{}} 0.6559*\\ $\uparrow$ 12.39\%\end{tabular}}     & \multicolumn{1}{c|}{\begin{tabular}[c]{@{}c@{}} 0.1601*\\ $\uparrow$ 17.03\%\end{tabular}}     &   \begin{tabular}[c]{@{}c@{}} 0.8261*\\ $\uparrow$ 15.98\%\end{tabular}      & \multicolumn{1}{c|}{\begin{tabular}[c]{@{}c@{}} \textbf{0.7563*}\\ $\uparrow$ \textbf{8.41\%}\end{tabular}}     & \multicolumn{1}{c|}{\begin{tabular}[c]{@{}c@{}} \textbf{0.7679*}\\ $\uparrow$ \textbf{8.66\%}\end{tabular}}     & \multicolumn{1}{c|}{\begin{tabular}[c]{@{}c@{}} 0.3625*\\ $\uparrow$ 5.69\%\end{tabular}}     &    \begin{tabular}[c]{@{}c@{}} 1.1302*\\ $\uparrow$ 19.86\%\end{tabular}     \\ \midrule
RERUM (CFR$_{mmd}$)                                       & \multicolumn{1}{c|}{\begin{tabular}[c]{@{}c@{}} 0.6242*\\ $\uparrow$ 3.55\%\end{tabular}}     & \multicolumn{1}{c|}{\begin{tabular}[c]{@{}c@{}} 0.6281*\\ $\uparrow$ 3.96\%\end{tabular}}     & \multicolumn{1}{c|}{\begin{tabular}[c]{@{}c@{}}\textbf{0.1581}*\\ $\uparrow$ \textbf{76.25\%} \end{tabular}}     &  \begin{tabular}[c]{@{}c@{}} 1.3015*\\ $\downarrow$ 3.78\% \end{tabular}       & \multicolumn{1}{c|}{\begin{tabular}[c]{@{}c@{}} 0.6374*\\ $\uparrow$ 8.81\% \end{tabular}}     & \multicolumn{1}{c|}{\begin{tabular}[c]{@{}c@{}} 0.6350*\\ $\uparrow$ 8.81\% \end{tabular}}     & \multicolumn{1}{c|}{\begin{tabular}[c]{@{}c@{}} \textbf{0.1692}*\\ $\uparrow$ \textbf{23.68\%} \end{tabular}}     &  \begin{tabular}[c]{@{}c@{}} 0.8252*\\ $\uparrow$ 15.85\% \end{tabular}       & \multicolumn{1}{c|}{\begin{tabular}[c]{@{}c@{}}0.7554*\\ $\uparrow$ 8.29\%\end{tabular}}     & \multicolumn{1}{c|}{\begin{tabular}[c]{@{}c@{}}0.7647*\\ $\uparrow$ 8.21\%\end{tabular}}     & \multicolumn{1}{c|}{\begin{tabular}[c]{@{}c@{}}0.3588*\\ $\uparrow$ 4.61\%\end{tabular}}     &   \begin{tabular}[c]{@{}c@{}}\textbf{1.2045*}\\ $\uparrow$ \textbf{27.74\%}\end{tabular}      \\ \midrule
RERUM (DragonNet)                                 & \multicolumn{1}{c|}{\begin{tabular}[c]{@{}c@{}} \textbf{0.6721*}\\ $\uparrow$ \textbf{11.50\%}\end{tabular}}    & \multicolumn{1}{c|}{\begin{tabular}[c]{@{}c@{}}\textbf{0.6753*}\\ $\uparrow$ \textbf{11.77\%} \end{tabular}}     & \multicolumn{1}{c|}{\begin{tabular}[c]{@{}c@{}}0.1434*\\ $\uparrow$ 59.87\% \end{tabular}}     & \begin{tabular}[c]{@{}c@{}} \textbf{1.5845}*\\ $\uparrow$ \textbf{17.14\%}\end{tabular}       & \multicolumn{1}{c|}{\begin{tabular}[c]{@{}c@{}} \textbf{0.6580*}\\ $\uparrow$ \textbf{12.33\%}\end{tabular}}     & \multicolumn{1}{c|}{\begin{tabular}[c]{@{}c@{}} \textbf{0.6566}*\\ $\uparrow$ \textbf{12.51\%}\end{tabular}}     & \multicolumn{1}{c|}{\begin{tabular}[c]{@{}c@{}} 0.1664*\\ $\uparrow$ 21.64\%\end{tabular}}     &  \begin{tabular}[c]{@{}c@{}} \textbf{0.9401*}\\ $\uparrow$ \textbf{31.98\%}\end{tabular}       & \multicolumn{1}{c|}{\begin{tabular}[c]{@{}c@{}}0.7176*\\ $\uparrow$ 2.87\%\end{tabular}}     & \multicolumn{1}{c|}{\begin{tabular}[c]{@{}c@{}}0.7359*\\ $\uparrow$ 4.13\%\end{tabular}}     & \multicolumn{1}{c|}{\begin{tabular}[c]{@{}c@{}}\textbf{0.3653*}\\ $\uparrow$ \textbf{6.50\%}\end{tabular}}     &     \begin{tabular}[c]{@{}c@{}}1.1016*\\ $\uparrow$ 16.83\%\end{tabular}    \\ \bottomrule
\end{tabular}}
\label{table:overall performance}
\end{table*}

\subsection{Experiment Setup}
\subsubsection{\textbf{Datasets}}

\begin{itemize}[leftmargin=*]
    \item \textbf{Hillstrom}~\citep{hillstrom2008minethatdata}: 
    This dataset is derived from an email merchandising campaign that involves 64,000 consumers. 
    The dataset contains three variables describing consumer activity in the following two weeks after the delivery of the email campaign: \textit{visit}, \textit{conversion}, and \textit{spend}. We select the \textit{spend} as the response label which indicates the actual dollar spent in the following two weeks, thus fitting our revenue uplift modeling problem setting.
    The original dataset has two types of treatment groups \textit{mens\_email}, \textit{womens\_email}, and a control group \textit{no\_email}. Following~\citep{betlei2021uplift}, we combine the \textit{mens\_email} treatment group and \textit{no email} control group as the \textbf{Hillstrom-Men} dataset and conduct the similar operation to the \textit{womens\_email} treatment group and \textit{no\_email} control group to obtain \textbf{Hillstrom-Women} dataset.
    
    \item \textbf{Product}: This is a product dataset that contains over 5,000,000 individuals, collected from the online mutual fund marketing scenario of Tencent FiT. There are more than 1,800 covariate features in the dataset and most of them are categorical. The response label is the amount that an individual pays to purchase funds. The treatment is an incentive coupon.
\end{itemize}
For these datasets, we perform a 60\%/10\%/30\% split to acquire the training set, validation set, and test set, respectively.

\subsubsection{\textbf{Baselines}}
We compare our RERUM method with the following baselines which are commonly adopted in uplift modeling: Causal Forest~\citep{davis2017using}, S-Learner~\citep{kunzel2019metalearners}, T-Learner~\citep{kunzel2019metalearners}, Transformed Response (TR)~\citep{athey2015machine}, TAR~\citep{shalit2017estimating}, CFR~\citep{shalit2017estimating}, StableCFR~\citep{wu2023stable}, CITE~\citep{li2022contrastive}, DragonNet~\citep{shi2019adapting} . Note that in the following experiments, we mainly take several prestigious deep learning-based uplift models~\citep{shalit2017estimating, shi2019adapting} as the backbones of our RERUM, considering their more competitive performance and wider application. We provide the detailed descriptions for these baselines in Appendix~\ref{sec:baselines}.

\subsubsection{\textbf{Evaluation Metrics}}
\label{sec:metric}
In our experiments, we adopt the following four metrics to evaluate the uplift ranking ability of different methods: Area Under the uplift curve (AUUC), Area Under the QINI Curve (AUQC), Kendall Rank Correlation Coefficient (KRCC), and LIFT@h ($h$ is set as 30). Following the previous work~\citep{devriendt2020learning}, we take the ``jointly, absolute'' version of uplift/Qini curves and further formulate the corresponding AUUC and AUQC metrics. For easier computation and fairer comparison, we utilize the \textit{normalized} AUUC and AUQC approximated over 100 buckets. The detailed evaluation metric introduction is provided in Appendix~\ref{sec: detailed metrics}.

\subsubsection{\textbf{Implementation Details}} We conduct each experiment with five different random seeds and take the average performance as the final result. We use the Adam optimizer with the initial learning rate as 0.001. The embedding size for each feature is set as 10 following the conventional design in industrial applications. We implement our method with PyTorch and run it on an NVIDIA Tesla V100 GPU with 32GB memory. Besides, we provide source codes in \textcolor{blue}{\url{https://github.com/BokwaiHo/revenue_uplift}}.

\vspace{-1mm}

\begin{table*}[!t]
\caption{Ablation study to our three modules under different base models in an incremental manner.}
\resizebox{0.99\textwidth}{!}{
\begin{tabular}{l|cccc|cccc|cccc}
\toprule
\multicolumn{1}{c|}{\multirow{2}{*}{Methods}} & \multicolumn{4}{c|}{Hillstrom-Men}                                                          & \multicolumn{4}{c|}{Hillstrom-Women}                                                        & \multicolumn{4}{c}{Product}                                                                \\ \cline{2-13} 
\multicolumn{1}{c|}{}                         & \multicolumn{1}{c|}{AUUC} & \multicolumn{1}{c|}{AUQC} & \multicolumn{1}{c|}{KRCC} & LIFT@30 & \multicolumn{1}{c|}{AUUC} & \multicolumn{1}{c|}{AUQC} & \multicolumn{1}{c|}{KRCC} & LIFT@30 & \multicolumn{1}{c|}{AUUC} & \multicolumn{1}{c|}{AUQC} & \multicolumn{1}{c|}{KRCC} & LIFT@30 \\ \midrule
TAR                                     & \multicolumn{1}{c|}{0.5652}     & \multicolumn{1}{c|}{0.5659}     & \multicolumn{1}{c|}{0.0780}     &    0.7974     & \multicolumn{1}{c|}{0.5739}     & \multicolumn{1}{c|}{0.5770}     & \multicolumn{1}{c|}{0.0586}     &   0.6366      & \multicolumn{1}{c|}{0.6493}     & \multicolumn{1}{c|}{0.6737}     & \multicolumn{1}{c|}{0.2986}     &    0.8918     \\ 
TAR+UR                                     & \multicolumn{1}{l|}{0.5958}     & \multicolumn{1}{c|}{0.5974}     & \multicolumn{1}{c|}{0.1083}     &  0.9098       & \multicolumn{1}{c|}{0.5980}     & \multicolumn{1}{c|}{0.5961}     & \multicolumn{1}{c|}{0.0947}     &    0.7067     & \multicolumn{1}{c|}{0.6725}     & \multicolumn{1}{c|}{0.7028}     & \multicolumn{1}{c|}{0.3091}     &  1.0485       \\ 
TAR+UR+RR                                     & \multicolumn{1}{c|}{0.6187}     & \multicolumn{1}{c|}{0.6199}     & \multicolumn{1}{c|}{0.1527}     & 1.1813        & \multicolumn{1}{c|}{0.6204}     & \multicolumn{1}{c|}{0.6197}     & \multicolumn{1}{c|}{0.1390}     &  0.7668      & \multicolumn{1}{c|}{0.6780}     & \multicolumn{1}{c|}{0.7071}     & \multicolumn{1}{c|}{0.3172}     &     1.0610    \\ 
RERUM (TAR)                                  & \multicolumn{1}{c|}{0.6299}     & \multicolumn{1}{c|}{0.6338}     & \multicolumn{1}{c|}{0.1617}     &   1.3477      & \multicolumn{1}{c|}{0.6360}     & \multicolumn{1}{c|}{0.6334}     & \multicolumn{1}{c|}{0.1806}     & 0.8755        & \multicolumn{1}{c|}{0.6863}     & \multicolumn{1}{c|}{0.7086}     & \multicolumn{1}{c|}{0.3269}     &  1.0247       \\ \midrule
CFR$_{wass}$                                      & \multicolumn{1}{c|}{0.5661}     & \multicolumn{1}{c|}{0.5676}     & \multicolumn{1}{c|}{0.0788}     &    0.8721     & \multicolumn{1}{c|}{0.5754}     & \multicolumn{1}{c|}{0.5762}     & \multicolumn{1}{c|}{0.0606}     & 0.6472        & \multicolumn{1}{c|}{0.6957}     & \multicolumn{1}{c|}{0.7018}     & \multicolumn{1}{c|}{0.2998}     &   0.8741      \\ 
CFR$_{wass}$+UR                                     & \multicolumn{1}{c|}{0.6027}     & \multicolumn{1}{c|}{0.6042}     & \multicolumn{1}{c|}{0.0844}     &  0.9020       & \multicolumn{1}{c|}{0.5973}     & \multicolumn{1}{c|}{0.5957}     & \multicolumn{1}{c|}{0.1247}     &  0.7261       & \multicolumn{1}{c|}{0.7150}     & \multicolumn{1}{c|}{0.7301}     & \multicolumn{1}{c|}{0.3451}     &   0.9721      \\ 
CFR$_{wass}$+UR+RR                                       & \multicolumn{1}{c|}{0.6150}     & \multicolumn{1}{c|}{0.6191}     & \multicolumn{1}{c|}{0.1349}     &  1.1506       & \multicolumn{1}{c|}{0.6144}     & \multicolumn{1}{c|}{0.6132}     & \multicolumn{1}{c|}{0.1490}     &  0.7442       & \multicolumn{1}{c|}{0.7384}     & \multicolumn{1}{c|}{0.7459}     & \multicolumn{1}{c|}{0.3507}     &    1.0291     \\ 
RERUM (CFR$_{wass}$)                                  & \multicolumn{1}{c|}{0.6474}     & \multicolumn{1}{c|}{0.6497}     & \multicolumn{1}{c|}{0.1382}     &  1.4602       & \multicolumn{1}{c|}{0.6568}     & \multicolumn{1}{c|}{0.6559}     & \multicolumn{1}{c|}{0.1601}     &   0.8261      & \multicolumn{1}{c|}{0.7563}     & \multicolumn{1}{c|}{0.7679}     & \multicolumn{1}{c|}{0.3625}     &    1.1302     \\ \midrule
CFR$_{mmd}$                                      & \multicolumn{1}{c|}{0.5760}     & \multicolumn{1}{c|}{0.5762}     & \multicolumn{1}{c|}{0.0747}     &   0.7351      & \multicolumn{1}{c|}{0.5836}     & \multicolumn{1}{c|}{0.5814}     & \multicolumn{1}{c|}{0.0788}     &  0.6206       & \multicolumn{1}{c|}{0.6933}     & \multicolumn{1}{c|}{0.7025}     & \multicolumn{1}{c|}{0.3430}     &     0.8929    \\ 
CFR$_{mmd}$+UR                                      & \multicolumn{1}{c|}{0.5981}     & \multicolumn{1}{c|}{0.5998}     & \multicolumn{1}{c|}{0.1160}     &   0.9163      & \multicolumn{1}{c|}{0.6081}     & \multicolumn{1}{c|}{0.5998}     & \multicolumn{1}{c|}{0.1301}     &  0.7193       & \multicolumn{1}{c|}{0.7127}     & \multicolumn{1}{c|}{0.7212}     & \multicolumn{1}{c|}{0.3697}     &     1.0126    \\ 
CFR$_{mmd}$+UR+RR                                      & \multicolumn{1}{c|}{0.6194}     & \multicolumn{1}{c|}{0.6206}     & \multicolumn{1}{c|}{0.1568}     &    1.1791     & \multicolumn{1}{c|}{0.6137}     & \multicolumn{1}{c|}{0.6114}     & \multicolumn{1}{c|}{0.1525}     &   0.7547      & \multicolumn{1}{c|}{0.7285}     & \multicolumn{1}{c|}{0.7357}     & \multicolumn{1}{c|}{0.3014}     &   1.0166      \\ 
RERUM (CFR$_{mmd}$)                                  & \multicolumn{1}{c|}{0.6242}     & \multicolumn{1}{c|}{0.6281}     & \multicolumn{1}{c|}{0.1581}     &    1.3015     & \multicolumn{1}{c|}{0.6374}     & \multicolumn{1}{c|}{0.6350}     & \multicolumn{1}{c|}{0.1692}     &  0.8252       & \multicolumn{1}{c|}{0.7554}     & \multicolumn{1}{c|}{0.7647}     & \multicolumn{1}{c|}{0.3588}     &    1.2045     \\ \midrule
DragonNet                                             & \multicolumn{1}{c|}{0.6028}     & \multicolumn{1}{c|}{0.6042}     & \multicolumn{1}{c|}{0.0897}     &   1.3526      & \multicolumn{1}{c|}{0.5858}     & \multicolumn{1}{c|}{0.5836}     & \multicolumn{1}{c|}{0.1368}     &    0.7123     & \multicolumn{1}{c|}{0.6347}     & \multicolumn{1}{c|}{0.6568}     & \multicolumn{1}{c|}{0.3253}     &   0.8148      \\ 
DragonNet+UR                                             & \multicolumn{1}{c|}{0.6534}     & \multicolumn{1}{c|}{0.6555}     & \multicolumn{1}{c|}{0.1155}     &   1.4343      & \multicolumn{1}{c|}{0.6184}     & \multicolumn{1}{c|}{0.6181}     & \multicolumn{1}{c|}{0.1695}     &  0.7891       & \multicolumn{1}{c|}{0.6623}     & \multicolumn{1}{c|}{0.6880}     & \multicolumn{1}{c|}{0.3224}     &   0.9277      \\ 
DragonNet+UR+RR                                         & \multicolumn{1}{c|}{0.6684}     & \multicolumn{1}{c|}{0.6713}     & \multicolumn{1}{c|}{0.1269}     &    1.5656     & \multicolumn{1}{c|}{0.6401}     & \multicolumn{1}{c|}{0.6394}     & \multicolumn{1}{c|}{0.1574}     &   0.9017      & \multicolumn{1}{c|}{0.6805}     & \multicolumn{1}{c|}{0.7040}     & \multicolumn{1}{c|}{0.3359}     &  1.0444       \\ 
RERUM (DragonNet)                                            & \multicolumn{1}{c|}{0.6721}     & \multicolumn{1}{c|}{0.6753}     & \multicolumn{1}{c|}{0.1434}     &      1.5845   & \multicolumn{1}{c|}{0.6580}     & \multicolumn{1}{c|}{0.6566}     & \multicolumn{1}{c|}{0.1664}     &  0.9401       & \multicolumn{1}{c|}{0.7176}     & \multicolumn{1}{c|}{0.7359}     & \multicolumn{1}{c|}{0.3653}     &    1.1016     \\ \bottomrule
\end{tabular}}
\label{table:ablation study}
\end{table*}

\subsection{Results and Analysis}
\subsubsection{\textbf{Overall Performance (RQ1)}}
We report the empirical results of our RERUM and baselines on three offline datasets in Table~\ref{table:overall performance}. First, we can easily observe that deep learning-based uplift models can reach better performance than tree/meta learner-based models under different settings, which validates the rationality that we take them as the base models of our RERUM. Second, we can observe that, though the uplift modeling capability differences among these base models, our four versions of RERUM with them can all achieve improvement over the best-performing baseline on most metrics and datasets. For example, our RERUM (DragonNet) improves the LIFT@30 metric by $21.98\%$ on average in three datasets. Third, comparing each version of RERUM with its corresponding base model, like RERUM (TAR) and TAR, our RERUM can consistently obtain significant performance gain. Taking the well-known AUUC metric as an example, the RERUM (TAR) brings the $11.45\%$, $10.82\%$, $5.70\%$ improvement on Hilstrom-Men, Hillstrom-Women, and Product datasets, respectively. All these results demonstrate the effectiveness and applicability of our RERUM for rankability-enhanced revenue uplift modeling.

\subsubsection{\textbf{Ablation Study (RQ2)}}
\label{sec:ablation study}
We conduct the ablation study by incrementally adding the uplift ranking learning module (UR), response ranking learning module (RR), and ZILN loss along with its corresponding response modeling framework, to the different base models in a sequential manner. The experimental results on three datasets are reported in Table~\ref{table:ablation study}. First, we can find that after the introduction of each module, the performance can all be strengthened to some extent on four ranking-related metrics, which demonstrates that our three contributions can all benefit the revenue uplift modeling. Second, comparing the performance of X+UR+RR and RERUM (X) (X indicates the base model), we can notice that the ZILN response regression module can effectively enhance the rankability of the uplift model, though it does not consider the uplift ranking explicitly. This also validates our motivation in Sec.~\ref{sec: uplift model framework}, that is, accurate revenue response modeling plays a crucial role in uplift modeling.

\subsubsection{\textbf{Hyperparameter Sensitivity Analysis (RQ3)}}
\begin{figure}[ht]
    \centering
    \includegraphics[width=0.48\textwidth]{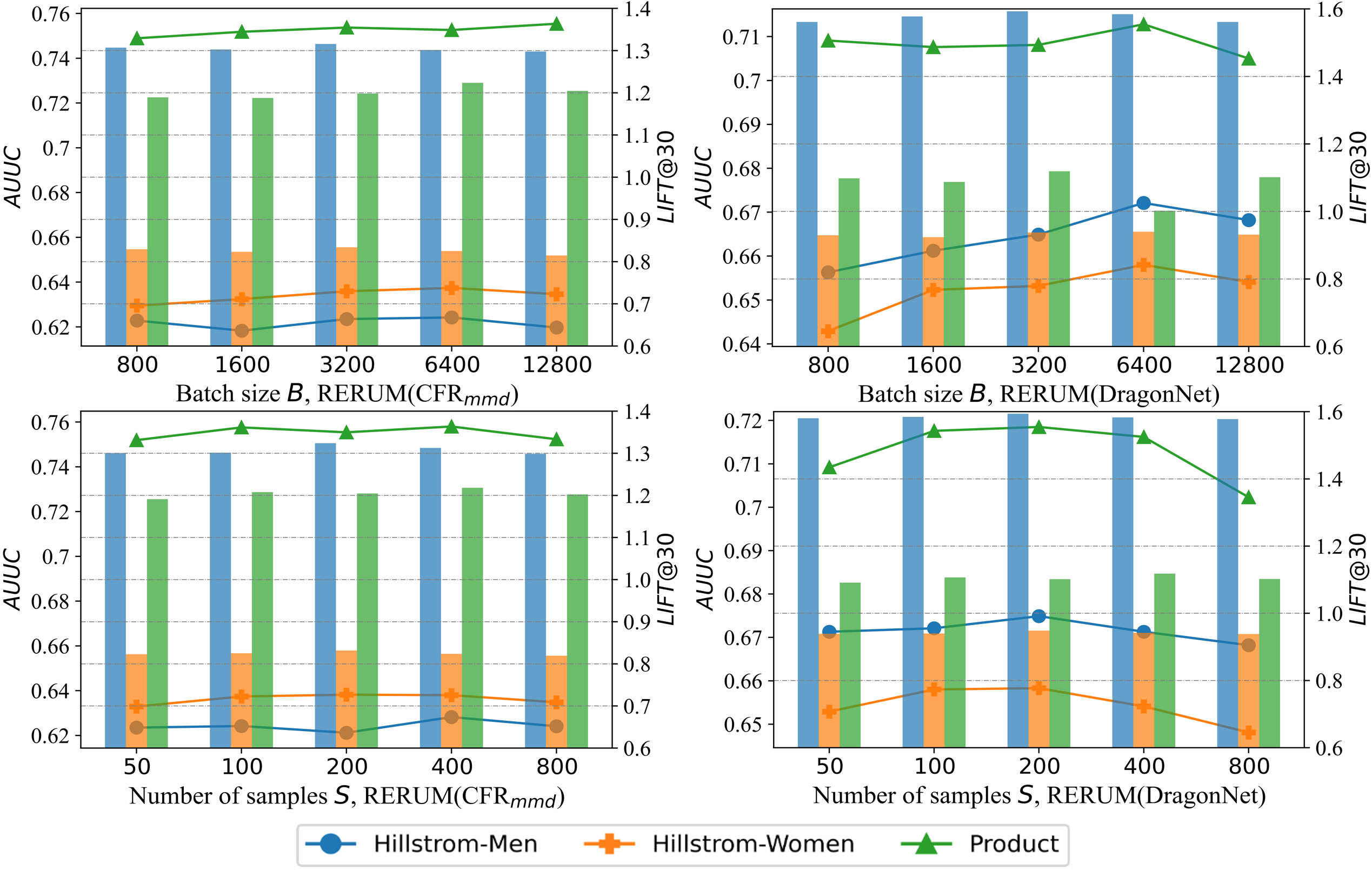}
    \caption{The hyperparameter sensitivity analysis results. The line chart corresponds to the AUUC metric and the column chart corresponds to the LIFT@30 metric.
    }
    \Description{The hyperparameter sensitivity analysis results. The line chart corresponds to the AUUC metric and the column chart corresponds to the LIFT@30 metric.}
    \label{fig: hyper}
    \vspace{-0.5cm}
\end{figure}

We conduct the sensitivity analysis on two essential hyperparameters: batch size $B$ which influences the estimation of $\mathcal{L}_{lu-rank}$ in Sec.~\ref{sec:url} and number of samples $S$ in Sec.~\ref{sec:rrl}. Due to the space limitation, we only present the results of RERUM (CFR$_{mmd}$) and RERUM (DragonNet) regarding AUUC and LIFT@30 metrics. The results are illustrated in Fig.~\ref{fig: hyper}. First, we can observe that both RERUM (CFR$_{mmd}$) and RERUM (DragonNet) are relatively robust to the hyperparameter selection, no matter of AUUC or LIFT@30 metric. Second, it can be noticed that under different hyperparameters, the performance of RERUM (DragonNet) fluctuates a little bit more than RERUM (CFR$_{mmd}$). This is due to the target regularization in DragonNet itself, which is also consistent with previous research.

\subsubsection{\textbf{Response Regression Performance Analysis (RQ4)}}
\begin{table}[!t]
\caption{Impacts of our proposed modules to response regression accuracy under different base models. Mean Absolute Percentage Error (MAPE) is taken as the metric.}
\begin{tabular}{l|c|c|c}
\toprule
\multicolumn{1}{c|}{\multirow{1}{*}{Methods}}                 & \multicolumn{1}{c|}{\begin{tabular}[c]{@{}c@{}}Hillstrom-\\Men\end{tabular}} & \multicolumn{1}{c|}{\begin{tabular}[c]{@{}c@{}}Hillstrom-\\Women\end{tabular}} & \multicolumn{1}{c}{Product} \\ \midrule
TAR                                     & \multicolumn{1}{c|}{0.2142}     & \multicolumn{1}{c|}{0.2087}     & \multicolumn{1}{c}{0.4429}        \\ 
TAR+ZILN                                     & \multicolumn{1}{c|}{0.0586}     & \multicolumn{1}{c|}{0.0514}     & \multicolumn{1}{c}{0.1378}     \\ 
TAR+ZILN+UR                                     & \multicolumn{1}{c|}{0.0614}     & \multicolumn{1}{c|}{0.0575}     & \multicolumn{1}{c}{0.1476}     \\ 
TAR+ZILN+RR                                     & \multicolumn{1}{c|}{0.0608}     & \multicolumn{1}{c|}{0.0550}     & \multicolumn{1}{c}{0.1423}     \\ 
RERUM (TAR)                                  & \multicolumn{1}{c|}{0.0667}     & \multicolumn{1}{c|}{0.0634}     & \multicolumn{1}{c}{0.1535}        \\ \midrule
CFR$_{wass}$                                      & \multicolumn{1}{c|}{0.2357}     & \multicolumn{1}{c|}{0.2024}     & \multicolumn{1}{c}{0.3966}       \\ 
CFR$_{wass}$+ZILN                                     & \multicolumn{1}{c|}{0.0651}     & \multicolumn{1}{c|}{0.0488}     & \multicolumn{1}{c}{0.1102}         \\ 
CFR$_{wass}$+ZILN+UR                                       & \multicolumn{1}{c|}{0.0685}     & \multicolumn{1}{c|}{0.0524}     & \multicolumn{1}{c}{0.1197}       \\ 
CFR$_{wass}$+ZILN+RR                                       & \multicolumn{1}{c|}{0.0677}     & \multicolumn{1}{c|}{0.0506}     & \multicolumn{1}{c}{0.1174}       \\ 
RERUM (CFR$_{wass}$)                                  & \multicolumn{1}{c|}{0.0741}     & \multicolumn{1}{c|}{0.0589}     & \multicolumn{1}{c}{0.1265}       \\ \midrule
CFR$_{mmd}$                                      & \multicolumn{1}{c|}{0.2456}     & \multicolumn{1}{c|}{0.2120}     & \multicolumn{1}{c}{0.4251}   \\ 
CFR$_{mmd}$+ZILN                                      & \multicolumn{1}{c|}{0.0712}     & \multicolumn{1}{c|}{0.0634}     & \multicolumn{1}{c}{0.1206}       \\ 
CFR$_{mmd}$+ZILN+UR                                      & \multicolumn{1}{c|}{0.0785}     & \multicolumn{1}{c|}{0.0697}     & \multicolumn{1}{c}{0.1385}       \\ 
CFR$_{mmd}$+ZILN+RR                                      & \multicolumn{1}{c|}{0.0764}     & \multicolumn{1}{c|}{0.0675}     & \multicolumn{1}{c}{0.1310}       \\ 
RERUM (CFR$_{mmd}$)                                  & \multicolumn{1}{c|}{0.0823}     & \multicolumn{1}{c|}{0.0731}     & \multicolumn{1}{c}{0.1436}     \\ \midrule
DragonNet                                             & \multicolumn{1}{c|}{0.1955}     & \multicolumn{1}{c|}{0.2037}     & \multicolumn{1}{c}{0.3754}       \\ 
DragonNet+ZILN                                             & \multicolumn{1}{c|}{0.0483}     & \multicolumn{1}{c|}{0.0528}     & \multicolumn{1}{c}{0.0986}       \\ 
DragonNet+ZILN+UR                                         & \multicolumn{1}{c|}{0.0545}     & \multicolumn{1}{c|}{0.0617}     & \multicolumn{1}{c}{0.1028}       \\ 
DragonNet+ZILN+RR                                         & \multicolumn{1}{c|}{0.0516}     & \multicolumn{1}{c|}{0.0574}     & \multicolumn{1}{c}{0.1001}       \\ 
RERUM (DragonNet)                                            & \multicolumn{1}{c|}{0.0628}     & \multicolumn{1}{c|}{0.0654}     & \multicolumn{1}{c}{0.1109}      \\ \bottomrule
\end{tabular}
\vspace{-4mm}
\label{table:regression performance analysis}
\end{table}

As emphasized in the Sec.~\ref{sec:intro}, the focus of this work is enhancing the uplift model's rankability and all three proposed modules are surrounding how to boost the model's capability to identify the most susceptible individuals to the treatment. Even utilizing ZILN instead of the conventional MSE to conduct the regression is also for improving ranking performance via accurate predicting the revenue response. Even though, investigating how such modules (especially UR and RR modules which are not specially designed for response regression) fare on the underlying response regression task can help provide the more comprehensive understanding to our methods. Therefore, we compare the performance of X (X indicates the base model), X+ZILN, X+ZILN+UR, RERUM(X), on the commonly used regression metric Mean Absolute Percentage Error (MAPE) of the response prediction (lower is better). Note that all base models X utilize the conventional MSE as the response regression loss according to their original papers. The detailed experimental results are provided in Table~\ref{table:regression performance analysis}.
From the Table~\ref{table:regression performance analysis}, we can find that X with ZILN as the regression loss can significantly outperform the X with the MSE as the regression loss on the MAPE metric. This demonstrates that the ZILN loss can effectively help predict the revenue responses more accurately, thus finally benefiting the uplift ranking. Besides, the introduction of UR and RR will influence the performance on MAPE metric a little bit. Because their main goal is to increase the uplift ranking accuray instead of the response regression. As long as they can facilitate the uplift ranking performance (shown in the Table~\ref{table:overall performance} of the paper), a small degree of decrease in response regression is still tolerable. Even though, the regression performance (MAPE) of X+ZILN+UR, X+ZILN+RR, RERUM(X) is still better than X with MSE as the regression loss.


\subsubsection{\textbf{Online Deployment (RQ5)}}
\label{sec:online}
\begin{figure}[ht]
    \centering
    \includegraphics[width=0.48\textwidth]{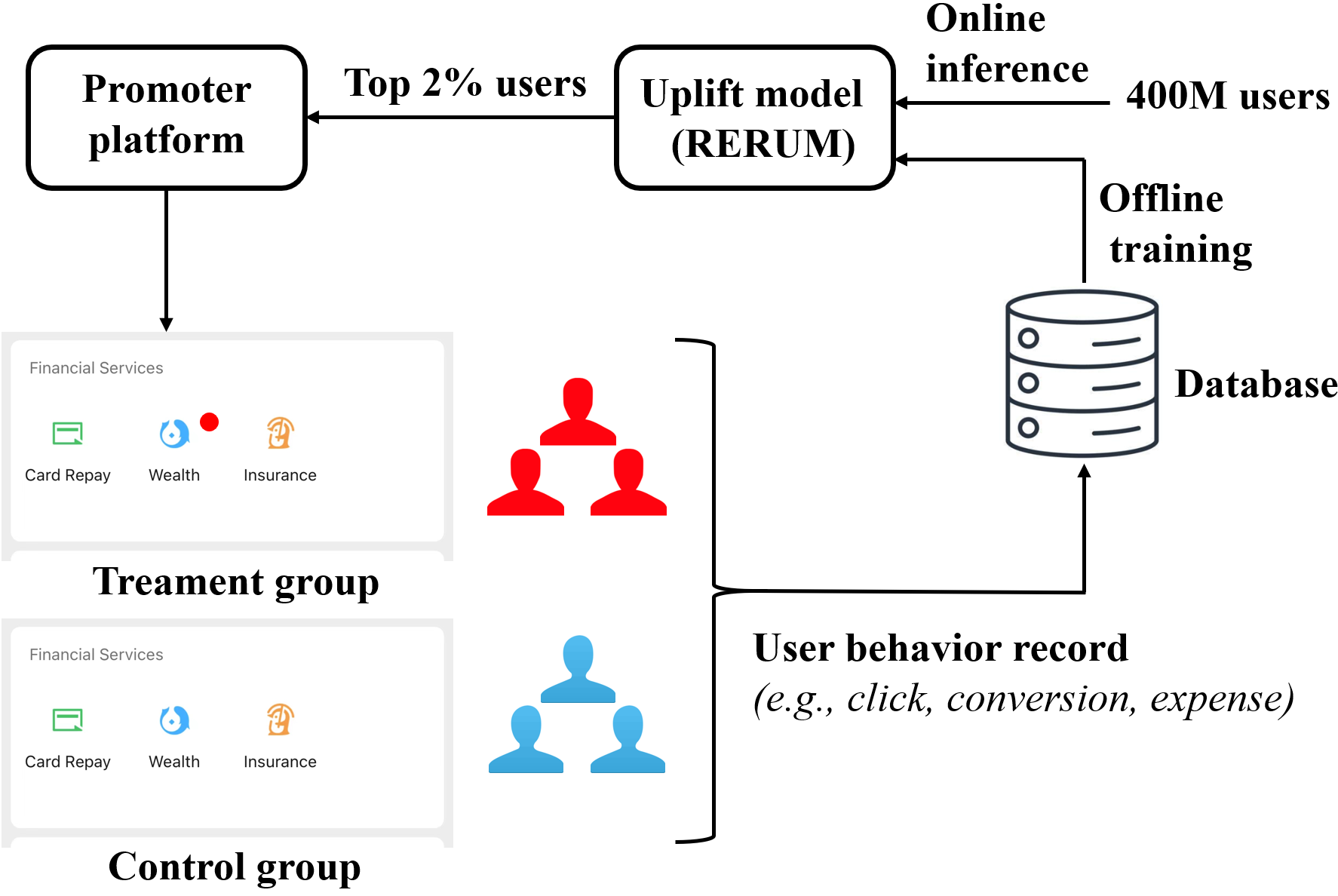}
    \caption{Overview of the online deployment platform.
    }
    \Description{Overview of the online deployment platform.}
    \label{fig: online platform}
    \vspace{-0.3cm}
\end{figure}

This is a mutual fund sales scenario with the notification redpoint as the treatment in the wealth management business of our deployment platform, Tencent FiT, one of the world's largest online fintech marketing platforms. 
Here, the response variable is the sales revenue. We first identify the top $2\%$ ranked individuals from the whole population of around 400 million users by the uplift model. Then, we randomly split such people into the treatment group and control group. The difference between the average response in such two groups in the following certain length of period (1 month), also known as the sales revenue LIFT@2, reflects the rankability of the model. The online deployment platform is illustrated in Fig.~\ref{fig: online platform}.
To enhance the reliability and validity, we conduct three times of such marketing campaigns to demonstrate the effectiveness of our RERUM.

\textit{Online Experimental Results Analysis.} From the results shown in Table~\ref{table:online performance}, we can observe that our RERUM achieves consistent improvement over the state-of-the-art (SOTA) online base model across three campaigns. The average improvement reaches $20.61\%$ on the sales revenue LIFT@2 metric. Besides, the RERUM brings 430 million dollar assets under management (AUM) gain each month. Such empirical results strongly demonstrate the effectiveness of our RERUM in real-world applications.

\begin{table}[]
\caption{Performance of SOTA online base model and our RERUM on sales revenue LIFT@2 in online deployment. }
\begin{tabular}{l|c|c|c}
\toprule
Methods   & Campaign 1 & Campaign 2 & Campaign 3 \\ \midrule
Base &     24566    &   30137  &  21971 \\ \midrule
RERUM(Base)      & 26826       &   41360  & 25361 \\  
Improvement       &   $\uparrow$ 9.20\%         &  $\uparrow$ 37.24\%      & $\uparrow$ 15.43\%   \\ \bottomrule
\end{tabular}
\label{table:online performance}
\vspace{-2mm}
\end{table}

%% file: conclusion.tex
\section{Conclusions and Future Work}
Revenue uplift modeling has been recognized as of great importance for online marketing. Especially, the uplift ranking within the whole population is the core to identifying the sensitive individuals to interventions. However, previous methods suffer from the low revenue response prediction precision and the neglection to the uplift ranking accuracy among individuals. In this paper, we have three contributions to address this problem: 1) modeling the response regression with a ZILN loss and a customized network framework that can be adapted with different base uplift models; 2) analyzing the uplift ranking errors and using two tighter response ranking losses to augment the vanilla regression term; 3) directly modeling the listwise uplift ranking among the whole population. The extensive experiments on both offline datasets and one of the world's largest online fintech marketing platforms demonstrate the effectiveness of our proposed method. In the future, we prepare to explore the revenue uplift modeling with multiple treatments and its application in cross-domain scenarios.

%% file: appendix.tex
\section{Main Notations}
To facilitate comprehension, we summarize the main notations used in this paper in Table~\ref{table:notation}.
\vspace{-1mm}
\begin{table}[h]
\caption{Major notations.}
\vspace{-2mm}
\begin{tabular}{c l}
\hline
$f$ & The uplift model\\
$\tau$ & The true uplift based on the potential responses\\
$\mathcal{D}^t$, $\mathcal{D}^c$ & \begin{tabular}[c]{@{}l@{}} The treatment group and control group\end{tabular} \\ 
$\mathbf{x}^t$, $\mathit{X}^t$     & \begin{tabular}[c]{@{}l@{}} The covariate vectors of users in treatment group and \\ their corresponding random variable\end{tabular} \\ 
$\mathbf{x}^c$, $\mathit{X}^c$ & \begin{tabular}[c]{@{}l@{}} The covariate vectors of users in control group and \\ their corresponding random variable \end{tabular} \\
$y^1$, $\mathit{Y}^1$     & \begin{tabular}[c]{@{}l@{}} The potential response instances under treatment and \\ their corresponding random variable\end{tabular} \\  
$y^0$, $\mathit{Y}^0$     & \begin{tabular}[c]{@{}l@{}} The potential response instances under control and \\ their corresponding random variable\end{tabular}\\ 
$\mathit{Y}, \mathit{X}, \mathit{T}$ & \begin{tabular}[c]{@{}l@{}} The general random variables that indicate the \\response, covariate, and treatment, respectively\end{tabular} \\
$\mathit{Y}_i, \mathit{X}_i, \mathit{T}_i$     & \begin{tabular}[c]{@{}l@{}} The random variables that indicate the individual \\ $i$'s response, covariate, and treatment, respectively\end{tabular}\\ \hline
\end{tabular}
\label{table:notation}
\end{table}
\vspace{-3mm}

\section{Descriptions to Baselines}
\label{sec:baselines}
We provide the detailed descriptions to all baselines as follows:
\begin{itemize}[leftmargin=*]
\item \textbf{Causal Forest}~\citep{davis2017using}: Causal Forest is a non-parametric Random Forest-based tree model that directly estimates the treatment effect, which is one of the most representative tree-based uplift models in many areas like economics and social science.
\item \textbf{S-Learner}~\citep{kunzel2019metalearners}: S-Learner is a kind of meta-learner method that uses a single estimator to estimate the response without giving the treatment a special role. The uplift is estimated by the difference between changed treatments with fixed covariates.
\item \textbf{T-Learner}~\citep{kunzel2019metalearners}: T-Learner is another type of meta-learner method. Different from S-Learner, T-Learner uses two estimators for the treatment group and control group respectively. 
\item \textbf{Transformed Response (TR)}~\citep{athey2015machine}: TR is a special technique that transforms the observed
response $Y$ to $Y^*$, such that the uplift equals the conditional expectation of the transformed response from the expectation perspective.
\item \textbf{TAR}~\citep{shalit2017estimating}: TAR is a commonly used deep learning-based uplift model. Compared with T-Learner, it omits the additional imputed treatment effects fitting sub-models but introduces the shared layers for treated and control response networks. The shared network parameters could help alleviate the sample imbalance issue between the treatment and control groups.
\item \textbf{CFR}~\citep{shalit2017estimating}: On the basis of TAR, CFR applies an additional loss to TAR, which forces the learned treated and control covariate distributions to be closer. We report the CFR performance using two distribution distance measurement loss functions, Wasserstein \citep{vallender1974calculation} (denoted as CFR$_{wass}$) and Maximum Mean Discrepancy (MMD) \citep{borgwardt2006integrating} (denoted as CFR$_{mmd}$).
\item \textbf{StableCFR}~\citep{wu2023stable}: StableCFR upsamples the underrepresented data with uniform sampling and balances covariates by using an epsilon-greedy matching mechanism to achieve higher uplift effect estimation accuracy for the underrepresented population.
\item \textbf{CITE}~\citep{li2022contrastive}: CITE is based on the contrastive task designed for causal inference, it fully exploits the self-supervision information hidden in data to achieve balanced and predictive representations while appropriately leveraging causal prior knowledge.
\item \textbf{DragonNet}~\citep{shi2019adapting}: Dragonnet exploits the sufficiency of the propensity score for estimation adjustment, and uses a regularization procedure based on non-parametric estimation theory which can guarantee desirable asymptotic properties.
\end{itemize}

\section{Descriptions to Evaluation Metrics}
\label{sec: detailed metrics}
We provide the detailed descriptions to our four uplift ranking evaluation metrics as follows:
\begin{itemize}[leftmargin=*]
\item \textbf{Area Under the Uplift Curve (AUUC)}: In uplift research, to evaluate the rankability of the uplift model $f$, one can first plot an uplift curve which ranks individual samples descendingly according to $predicted$ uplift $\hat{\tau}$ (in X-axis) and cumulatively sums the $observed$ uplift (in Y-axis)~\citep{betlei2021uplift}. The AUUC is then the area under this curve. There are actually multiple variants of uplift curves proposed in the recent literature. Their differences mainly lie in 1) if ranking the data separately per group or jointly over all data, and 2) if expressing volumes in absolute or relative numbers~\citep{devriendt2020learning}. In this work, we take the ``jointly, absolute'' uplift curves in~\citep{devriendt2020learning} and formulate the AUUC. First, we denote the total number of treated and control instances, among the top-$k$ individuals $\pi(\mathcal{D}, k)$ ranked by uplift model $f$ over the whole dataset $\mathcal{D}$ as
\begin{equation}
\begin{aligned}
    N^T_{\pi} (\mathcal{D}, k) &= \underset{(\mathbf{x}_i, t_i, y_i) \in \pi(\mathcal{D}, k)}{\sum} \mathbb{I} (t_i =1), \\
    N^C_{\pi} (\mathcal{D}, k) &= \underset{(\mathbf{x}_i, t_i, y_i) \in \pi(\mathcal{D}, k)}{\sum} \mathbb{I} (t_i =0), 
\end{aligned}
\label{eq:nt/c}
\end{equation}
where the $\mathbb{I} (\cdot)$ is the indicator function. Then, the response summation of treated and control individuals in $\pi(\mathcal{D}, k)$ are: 
\begin{equation}
\begin{aligned}
    R^T_{\pi} (\mathcal{D}, k) &= \underset{(\mathbf{x}_i, t_i, y_i) \in \pi(\mathcal{D}, k)}{\sum} y_i\mathbb{I} (t_i =1), \\
    R^C_{\pi} (\mathcal{D}, k) &= \underset{(\mathbf{x}_i, t_i, y_i) \in \pi(\mathcal{D}, k)}{\sum} y_i\mathbb{I} (t_i =0), 
\end{aligned}
\label{eq:rt/c}
\end{equation}
where $y_i$ is the observed response. Thus, the values for the uplift curve can be obtained with:
\begin{equation}
\begin{aligned}
     V_u(f, k) = (\frac{R^T_{\pi} (\mathcal{D}, k)}{N^T_{\pi} (\mathcal{D}, k)} - \frac{R^C_{\pi} (\mathcal{D}, k)}{N^C_{\pi} (\mathcal{D}, k)}) * (N^T_{\pi} (\mathcal{D}, k) + N^C_{\pi} (\mathcal{D}, k)).
\end{aligned}
\label{eq:uplift curve}
\end{equation}
Furthermore, the AUUC is formulated with:
\begin{equation}
    AUUC(f, k) = \int_0^1 V_u(f, x) dx = \sum_{k=1}^n V_u(f,k) \approx \sum_{p=1}^{100} V_u(f,\frac{p}{100}).
\label{eq:auuc}
\end{equation}
For the separate setting, we can make the above approximation over 100 buckets to estimate the AUUC. In this paper, we utilize the \textit{normalized} AUUC for the fairer comparison.
\item \textbf{Area Under the QINI Curve (AUQC)}: Similar to the above AUUC, we follow ~\citep{devriendt2020learning} and take the ``jointly, absolute'' Qini curve to formulate AUQC metric. Based on the previous definition in Eq.~\ref{eq:nt/c} and ~\ref{eq:rt/c}, we define the values for the Qini curve as:
\begin{equation}
\begin{aligned}
     V_q(f, k) = R^T_{\pi} (\mathcal{D}, k) - R^C_{\pi} (\mathcal{D}, k) *\frac{N^T_{\pi} (\mathcal{D}, k)}{N^C_{\pi} (\mathcal{D}, k)}. 
\end{aligned}
\label{eq:qini curve}
\end{equation}
Thus, the AUQC can be obtained as:
\begin{equation}
    AUQC(f, k) = \int_0^1 V_q(f, x) dx = \sum_{k=1}^n V_q(f,k) \approx \sum_{p=1}^{100} V_q(f,\frac{p}{100}).
\label{eq:auqc}
\end{equation}
We use the \textit{normalized} AUQC to compare different methods.

\item \textbf{Kendall Rank Correlation Coefficient (KRCC)}: The KRCC~\citep{abdi2007kendall} measures the similarity between the rank by predicted uplift scores and the rank by the approximated true uplift scores for all individuals. To facilitate the computation, we split the whole population into 100 buckets and use the treatment and control group data in each bucket to approximate the true uplift effect.

\item \textbf{LIFT@h}: This metric measures the difference between the mean response of treated individuals and that of controlled individuals in top $h$ percentile of all individuals ranked by the uplift model. It has been widely employed in many industrial scenarios, because it can explicitly reflect the model's rankability, especially for a certain proportion of targeted people. Here, we take $h$ as 30.
\end{itemize}

\begin{table*}[!t]
\caption{Ablation study to our three modules under different base models by incrementally adding ZILN regression module (ZILN), uplift ranking learning module (UR), and response ranking learning module (RR).}
\resizebox{\textwidth}{!}{
\begin{tabular}{l|cccc|cccc|cccc}
\toprule
\multicolumn{1}{c|}{\multirow{2}{*}{Methods}} & \multicolumn{4}{c|}{Hillstrom-Men}                                                          & \multicolumn{4}{c|}{Hillstrom-Women}                                                        & \multicolumn{4}{c}{Product}                                                                \\ \cline{2-13} 
\multicolumn{1}{c|}{}                         & \multicolumn{1}{c|}{AUUC} & \multicolumn{1}{c|}{AUQC} & \multicolumn{1}{c|}{KRCC} & LIFT@30 & \multicolumn{1}{c|}{AUUC} & \multicolumn{1}{c|}{AUQC} & \multicolumn{1}{c|}{KRCC} & LIFT@30 & \multicolumn{1}{c|}{AUUC} & \multicolumn{1}{c|}{AUQC} & \multicolumn{1}{c|}{KRCC} & LIFT@30 \\ \midrule
TAR                                     & \multicolumn{1}{c|}{0.5652}     & \multicolumn{1}{c|}{0.5659}     & \multicolumn{1}{c|}{0.0780}     &    0.7974     & \multicolumn{1}{c|}{0.5739}     & \multicolumn{1}{c|}{0.5770}     & \multicolumn{1}{c|}{0.0586}     &   0.6366      & \multicolumn{1}{c|}{0.6493}     & \multicolumn{1}{c|}{0.6737}     & \multicolumn{1}{c|}{0.2986}     &    0.8918     \\ 
TAR+ZILN                                     & \multicolumn{1}{l|}{0.5741}     & \multicolumn{1}{c|}{0.5784}     & \multicolumn{1}{c|}{0.0912}     &  0.8216       & \multicolumn{1}{c|}{0.5972}     & \multicolumn{1}{c|}{0.5946}     & \multicolumn{1}{c|}{0.0831}     &    0.6687     & \multicolumn{1}{c|}{0.6679}     & \multicolumn{1}{c|}{0.6910}     & \multicolumn{1}{c|}{0.3105}     &  0.9211       \\ 
TAR+ZILN+UR                                     & \multicolumn{1}{c|}{0.6028}     & \multicolumn{1}{c|}{0.6131}     & \multicolumn{1}{c|}{0.1407}     & 1.1625        & \multicolumn{1}{c|}{0.6206}     & \multicolumn{1}{c|}{0.6218}     & \multicolumn{1}{c|}{0.1577}     &  0.7384      & \multicolumn{1}{c|}{0.6813}     & \multicolumn{1}{c|}{0.7022}     & \multicolumn{1}{c|}{0.3209}     &     0.9885    \\ 
RERUM (TAR)                                  & \multicolumn{1}{c|}{0.6299}     & \multicolumn{1}{c|}{0.6338}     & \multicolumn{1}{c|}{0.1617}     &   1.3477      & \multicolumn{1}{c|}{0.6360}     & \multicolumn{1}{c|}{0.6334}     & \multicolumn{1}{c|}{0.1806}     & 0.8755        & \multicolumn{1}{c|}{0.6863}     & \multicolumn{1}{c|}{0.7086}     & \multicolumn{1}{c|}{0.3269}     &  1.0247       \\ \midrule
CFR$_{wass}$                                      & \multicolumn{1}{c|}{0.5661}     & \multicolumn{1}{c|}{0.5676}     & \multicolumn{1}{c|}{0.0788}     &    0.8721     & \multicolumn{1}{c|}{0.5754}     & \multicolumn{1}{c|}{0.5762}     & \multicolumn{1}{c|}{0.0606}     & 0.6472        & \multicolumn{1}{c|}{0.6957}     & \multicolumn{1}{c|}{0.7018}     & \multicolumn{1}{c|}{0.2998}     &   0.8741      \\ 
CFR$_{wass}$+ZILN                                     & \multicolumn{1}{c|}{0.5816}     & \multicolumn{1}{c|}{0.5924}     & \multicolumn{1}{c|}{0.1052}     &  0.9002       & \multicolumn{1}{c|}{0.5904}     & \multicolumn{1}{c|}{0.5933}     & \multicolumn{1}{c|}{0.0782}     &  0.6671       & \multicolumn{1}{c|}{0.7025}     & \multicolumn{1}{c|}{0.7068}     & \multicolumn{1}{c|}{0.3112}     &   0.8946      \\ 
CFR$_{wass}$+ZILN+UR                                       & \multicolumn{1}{c|}{0.6214}     & \multicolumn{1}{c|}{0.6282}     & \multicolumn{1}{c|}{0.1237}     &  1.2841       & \multicolumn{1}{c|}{0.6335}     & \multicolumn{1}{c|}{0.6361}     & \multicolumn{1}{c|}{0.1323}     &  0.7835       & \multicolumn{1}{c|}{0.7364}     & \multicolumn{1}{c|}{0.7438}     & \multicolumn{1}{c|}{0.3442}     &    1.0824     \\ 
RERUM (CFR$_{wass}$)                                  & \multicolumn{1}{c|}{0.6474}     & \multicolumn{1}{c|}{0.6497}     & \multicolumn{1}{c|}{0.1382}     &  1.4602       & \multicolumn{1}{c|}{0.6568}     & \multicolumn{1}{c|}{0.6559}     & \multicolumn{1}{c|}{0.1601}     &   0.8261      & \multicolumn{1}{c|}{0.7563}     & \multicolumn{1}{c|}{0.7679}     & \multicolumn{1}{c|}{0.3625}     &    1.1302     \\ \midrule
CFR$_{mmd}$                                      & \multicolumn{1}{c|}{0.5760}     & \multicolumn{1}{c|}{0.5762}     & \multicolumn{1}{c|}{0.0747}     &   0.7351      & \multicolumn{1}{c|}{0.5836}     & \multicolumn{1}{c|}{0.5814}     & \multicolumn{1}{c|}{0.0788}     &  0.6206       & \multicolumn{1}{c|}{0.6933}     & \multicolumn{1}{c|}{0.7025}     & \multicolumn{1}{c|}{0.3430}     &     0.8929    \\ 
CFR$_{mmd}$+ZILN                                      & \multicolumn{1}{c|}{0.5942}     & \multicolumn{1}{c|}{0.5964}     & \multicolumn{1}{c|}{0.1138}     &   0.7590      & \multicolumn{1}{c|}{0.5933}     & \multicolumn{1}{c|}{0.5928}     & \multicolumn{1}{c|}{0.0923}     &  0.6418       & \multicolumn{1}{c|}{0.7023}     & \multicolumn{1}{c|}{0.7186}     & \multicolumn{1}{c|}{0.3491}     &     0.9112    \\ 
CFR$_{mmd}$+ZILN+UR                                      & \multicolumn{1}{c|}{0.6125}     & \multicolumn{1}{c|}{0.6153}     & \multicolumn{1}{c|}{0.1406}     &    1.1247     & \multicolumn{1}{c|}{0.6213}     & \multicolumn{1}{c|}{0.6204}     & \multicolumn{1}{c|}{0.1429}     &   0.7846      & \multicolumn{1}{c|}{0.7343}     & \multicolumn{1}{c|}{0.7425}     & \multicolumn{1}{c|}{0.3527}     &   1.1136      \\ 
RERUM (CFR$_{mmd}$)                                  & \multicolumn{1}{c|}{0.6242}     & \multicolumn{1}{c|}{0.6281}     & \multicolumn{1}{c|}{0.1581}     &    1.3015     & \multicolumn{1}{c|}{0.6374}     & \multicolumn{1}{c|}{0.6350}     & \multicolumn{1}{c|}{0.1692}     &  0.8252       & \multicolumn{1}{c|}{0.7554}     & \multicolumn{1}{c|}{0.7647}     & \multicolumn{1}{c|}{0.3588}     &    1.2045     \\ \midrule
DragonNet                                             & \multicolumn{1}{c|}{0.6028}     & \multicolumn{1}{c|}{0.6042}     & \multicolumn{1}{c|}{0.0897}     &   1.3526      & \multicolumn{1}{c|}{0.5858}     & \multicolumn{1}{c|}{0.5836}     & \multicolumn{1}{c|}{0.1368}     &    0.7123     & \multicolumn{1}{c|}{0.6347}     & \multicolumn{1}{c|}{0.6568}     & \multicolumn{1}{c|}{0.3253}     &   0.8148      \\ 
DragonNet+ZILN                                             & \multicolumn{1}{c|}{0.6204}     & \multicolumn{1}{c|}{0.6227}     & \multicolumn{1}{c|}{0.1235}     &   1.4167      & \multicolumn{1}{c|}{0.6124}     & \multicolumn{1}{c|}{0.6140}     & \multicolumn{1}{c|}{0.1435}     &  0.7482       & \multicolumn{1}{c|}{0.6627}     & \multicolumn{1}{c|}{0.6835}     & \multicolumn{1}{c|}{0.3349}     &   0.8582      \\ 
DragonNet+ZILN+UR                                         & \multicolumn{1}{c|}{0.6587}     & \multicolumn{1}{c|}{0.6603}     & \multicolumn{1}{c|}{0.1395}     &    1.5274     & \multicolumn{1}{c|}{0.6402}     & \multicolumn{1}{c|}{0.6419}     & \multicolumn{1}{c|}{0.1581}     &   0.8826      & \multicolumn{1}{c|}{0.6935}     & \multicolumn{1}{c|}{0.7148}     & \multicolumn{1}{c|}{0.3564}     &  1.0274       \\ 
RERUM (DragonNet)                                            & \multicolumn{1}{c|}{0.6721}     & \multicolumn{1}{c|}{0.6753}     & \multicolumn{1}{c|}{0.1434}     &      1.5845   & \multicolumn{1}{c|}{0.6580}     & \multicolumn{1}{c|}{0.6566}     & \multicolumn{1}{c|}{0.1664}     &  0.9401       & \multicolumn{1}{c|}{0.7176}     & \multicolumn{1}{c|}{0.7359}     & \multicolumn{1}{c|}{0.3653}     &    1.1016     \\ \bottomrule
\end{tabular}}
\label{table:ablation study 2}
\end{table*}

\begin{table*}[!t]
\caption{Ablation study to within-group response ranking loss and cross-group response ranking loss, respectively.}
\resizebox{\textwidth}{!}{
\begin{tabular}{l|cccc|cccc|cccc}
\toprule
\multicolumn{1}{c|}{\multirow{2}{*}{Methods}} & \multicolumn{4}{c|}{Hillstrom-Men}                                                          & \multicolumn{4}{c|}{Hillstrom-Women}                                                        & \multicolumn{4}{c}{Product}                                                                \\ \cline{2-13} 
\multicolumn{1}{c|}{}                         & \multicolumn{1}{c|}{AUUC} & \multicolumn{1}{c|}{AUQC} & \multicolumn{1}{c|}{KRCC} & LIFT@30 & \multicolumn{1}{c|}{AUUC} & \multicolumn{1}{c|}{AUQC} & \multicolumn{1}{c|}{KRCC} & LIFT@30 & \multicolumn{1}{c|}{AUUC} & \multicolumn{1}{c|}{AUQC} & \multicolumn{1}{c|}{KRCC} & LIFT@30 \\ \midrule
RERUM (TAR) w/o wr-rank                                     & \multicolumn{1}{l|}{0.6203}     & \multicolumn{1}{c|}{0.6241}     & \multicolumn{1}{c|}{0.1293}     &  1.1927       & \multicolumn{1}{c|}{0.6214}     & \multicolumn{1}{c|}{0.6207}     & \multicolumn{1}{c|}{0.1561}     &    0.8382     & \multicolumn{1}{c|}{0.6825}     & \multicolumn{1}{c|}{0.7030}     & \multicolumn{1}{c|}{0.3184}     &  1.0032       \\ 
RERUM (TAR) w/o cr-rank                                    & \multicolumn{1}{l|}{0.6228}     & \multicolumn{1}{c|}{0.6272}     & \multicolumn{1}{c|}{0.1385}     &  1.2364       & \multicolumn{1}{c|}{0.6256}     & \multicolumn{1}{c|}{0.6243}     & \multicolumn{1}{c|}{0.1620}     &    0.8541     & \multicolumn{1}{c|}{0.6839}     & \multicolumn{1}{c|}{0.7051}     & \multicolumn{1}{c|}{0.3207}     &  1.0125       \\ 
RERUM (TAR)                                  & \multicolumn{1}{c|}{0.6299}     & \multicolumn{1}{c|}{0.6338}     & \multicolumn{1}{c|}{0.1617}     &   1.3477      & \multicolumn{1}{c|}{0.6360}     & \multicolumn{1}{c|}{0.6334}     & \multicolumn{1}{c|}{0.1806}     & 0.8755        & \multicolumn{1}{c|}{0.6863}     & \multicolumn{1}{c|}{0.7086}     & \multicolumn{1}{c|}{0.3269}     &  1.0247       \\ \midrule
RERUM (CFR$_{wass}$) w/o wr-rank                                     & \multicolumn{1}{c|}{0.6384}     & \multicolumn{1}{c|}{0.6401}     & \multicolumn{1}{c|}{0.1185}     &  1.3425       & \multicolumn{1}{c|}{0.6437}     & \multicolumn{1}{c|}{0.6430}     & \multicolumn{1}{c|}{0.1477}     &  0.8133       & \multicolumn{1}{c|}{0.7489}     & \multicolumn{1}{c|}{0.7576}     & \multicolumn{1}{c|}{0.3539}     &   1.0986      \\ 
RERUM (CFR$_{wass}$) w/o cr-rank                                       & \multicolumn{1}{c|}{0.6405}     & \multicolumn{1}{c|}{0.6422}     & \multicolumn{1}{c|}{0.1236}     &  1.3861       & \multicolumn{1}{c|}{0.6482}     & \multicolumn{1}{c|}{0.6489}     & \multicolumn{1}{c|}{0.1542}     &  0.8192       & \multicolumn{1}{c|}{0.7521}     & \multicolumn{1}{c|}{0.7605}     & \multicolumn{1}{c|}{0.3574}     &    1.1133     \\ 
RERUM (CFR$_{wass}$)                                  & \multicolumn{1}{c|}{0.6474}     & \multicolumn{1}{c|}{0.6497}     & \multicolumn{1}{c|}{0.1382}     &  1.4602       & \multicolumn{1}{c|}{0.6568}     & \multicolumn{1}{c|}{0.6559}     & \multicolumn{1}{c|}{0.1601}     &   0.8261      & \multicolumn{1}{c|}{0.7563}     & \multicolumn{1}{c|}{0.7679}     & \multicolumn{1}{c|}{0.3625}     &    1.1302     \\ \midrule
RERUM (CFR$_{mmd}$) w/o wr-rank                                      & \multicolumn{1}{c|}{0.6188}     & \multicolumn{1}{c|}{0.6219}     & \multicolumn{1}{c|}{0.1395}     &   1.1721      & \multicolumn{1}{c|}{0.6288}     & \multicolumn{1}{c|}{0.6274}     & \multicolumn{1}{c|}{0.1583}     &  0.8160       & \multicolumn{1}{c|}{0.7448}     & \multicolumn{1}{c|}{0.7539}     & \multicolumn{1}{c|}{0.3487}     &     1.1208    \\ 
RERUM (CFR$_{mmd}$) w/o cr-rank                                      & \multicolumn{1}{c|}{0.6215}     & \multicolumn{1}{c|}{0.6240}     & \multicolumn{1}{c|}{0.1467}     &    1.2258     & \multicolumn{1}{c|}{0.6313}     & \multicolumn{1}{c|}{0.6299}     & \multicolumn{1}{c|}{0.1616}     &   0.8192      & \multicolumn{1}{c|}{0.7483}     & \multicolumn{1}{c|}{0.7565}     & \multicolumn{1}{c|}{0.3512}     &   1.1519      \\ 
RERUM (CFR$_{mmd}$)                                  & \multicolumn{1}{c|}{0.6242}     & \multicolumn{1}{c|}{0.6281}     & \multicolumn{1}{c|}{0.1581}     &    1.3015     & \multicolumn{1}{c|}{0.6374}     & \multicolumn{1}{c|}{0.6350}     & \multicolumn{1}{c|}{0.1692}     &  0.8252       & \multicolumn{1}{c|}{0.7554}     & \multicolumn{1}{c|}{0.7647}     & \multicolumn{1}{c|}{0.3588}     &    1.2045     \\ \midrule
RERUM (DragonNet) w/o wr-rank                                            & \multicolumn{1}{c|}{0.6639}     & \multicolumn{1}{c|}{0.6672}     & \multicolumn{1}{c|}{0.1298}     &   1.4763      & \multicolumn{1}{c|}{0.6465}     & \multicolumn{1}{c|}{0.6482}     & \multicolumn{1}{c|}{0.1545}     &  0.9287       & \multicolumn{1}{c|}{0.7052}     & \multicolumn{1}{c|}{0.7274}     & \multicolumn{1}{c|}{0.3561}     &   0.9852      \\ 
RERUM (DragonNet) w/o cr-rank                                          & \multicolumn{1}{c|}{0.6673}     & \multicolumn{1}{c|}{0.6701}     & \multicolumn{1}{c|}{0.1365}     &    1.5272     & \multicolumn{1}{c|}{0.6519}     & \multicolumn{1}{c|}{0.6528}     & \multicolumn{1}{c|}{0.1617}     &   0.9340      & \multicolumn{1}{c|}{0.7089}     & \multicolumn{1}{c|}{0.7302}     & \multicolumn{1}{c|}{0.3596}     &  1.0324       \\ 
RERUM (DragonNet)                                            & \multicolumn{1}{c|}{0.6721}     & \multicolumn{1}{c|}{0.6753}     & \multicolumn{1}{c|}{0.1434}     &      1.5845   & \multicolumn{1}{c|}{0.6580}     & \multicolumn{1}{c|}{0.6566}     & \multicolumn{1}{c|}{0.1664}     &  0.9401       & \multicolumn{1}{c|}{0.7176}     & \multicolumn{1}{c|}{0.7359}     & \multicolumn{1}{c|}{0.3653}     &    1.1016     \\ \bottomrule
\end{tabular}}
\label{table:ablation study 3}
\end{table*}

\section{Supplementary Ablation Study}
\label{sec:supp ablation}
In addition to the ablation study conducted in Sec.~\ref{sec:ablation study}, we also adjust the module stacking order to further highlight the effect of each module to the revenue uplift modeling. In detail, we incrementally add the ZILN regression module (ZILN), uplift ranking learning module (UR), and response ranking learning module (RR) to the vanilla base model X. The corresponding experimental results are illustrated in Table~\ref{table:ablation study 2}. First, it is evident that X+ZILN can achieve better performance than vinalla X under almost all ranking-related metrics and datasets, which implies that ZILN module can indeed improve the rankability of uplift models. This is actually consistent with our discussion in Sec.~\ref{sec: uplift model framework}. Second, we can also notice that integrating each module into the overall framework can respectively bring about improvement in ranking performance, which further verifies their effectiveness.

To demonstrate that both within-group response ranking loss (wr-rank) and cross-group ranking loss (cr-rank) in Sec.~\ref{sec:rrl} contribute positively to the final uplift ranking performance, we specially remove them from the overall scheme respectively and observe the corresponding results. The complete version of results are present in Table~\ref{table:ablation study 3}. From this table, we can first observe that both RERUM(X) w/o wr-rank and RERUM(X) w/o cr-rank achieve lower ranking performance than RERUM(X), which demonstrates that both wr-rank and cr-rank contribute positively to the final performance, i.e., both wr-rank and cr-rank are necessary. Besides, we can find that the performance drop of removing wr-rank is a little more than that brought by removing cr-ranking, which implies that the effect of cr-rank is a little bit weaker than wr-rank.
